\documentclass[runningheads]{llncs}

\usepackage{eccv}

\usepackage{eccvabbrv}

\usepackage{graphicx}
\usepackage{booktabs}
\usepackage{dsfont}
\usepackage{multirow}
\usepackage{wrapfig}
\newcommand{\xhdr}[1]{ \noindent {\textbf{#1}}}
\newcommand{\eqnsm}[2]{\begin{equation}\label{eq:#1}#2\end{equation}}
\newcommand{\TP}{\mathit{TP}}
\newcommand{\FN}{\mathit{FN}}
\newcommand{\FP}{\mathit{FP}}

\usepackage[accsupp]{axessibility}  

\usepackage[pagebackref,breaklinks,colorlinks,citecolor=eccvblue]{hyperref}

\begin{document}

\title{3D Open-Vocabulary Panoptic Segmentation with\\2D-3D Vision-Language Distillation} 

\titlerunning{3D Open-Vocabulary Panoptic Segmentation}

\author{Zihao Xiao\inst{1*} \and Longlong Jing\inst{2} \and Shangxuan Wu\inst{2} \and Alex Zihao  Zhu\inst{2} \and Jingwei Ji\inst{2} \and Chiyu Max Jiang\inst{2} \and Wei-Chih Hung\inst{2} \and  Thomas Funkhouser\inst{3} \and  Weicheng Kuo\inst{4} \and Anelia Angelova\inst{4} \and Yin Zhou\inst{2}  \and Shiwei Sheng\inst{2*} \\
}

\authorrunning{Zihao Xiao \emph{et al.}}

\institute{$^{1}$ Johns Hopkins University, $^{2}$ Waymo,  $^{3}$ Google Research, $^{4}$ Google DeepMind}

\maketitle
\begin{abstract}
{
3D panoptic segmentation is a challenging perception task, especially in autonomous driving. It aims to predict both semantic and instance annotations for 3D points in a scene. Although prior 3D panoptic segmentation approaches have achieved great performance on closed-set benchmarks, generalizing these approaches to unseen \textit{things} and unseen \textit{stuff} categories remains an open problem. For unseen object categories, 2D open-vocabulary segmentation has achieved promising results that solely rely on frozen CLIP backbones and ensembling multiple classification outputs. However, we find that simply extending these 2D models to 3D does not guarantee good performance due to poor per-mask classification quality, especially for novel \textit{stuff} categories. In this paper, we propose the first method to tackle 3D open-vocabulary panoptic segmentation. Our model takes advantage of the fusion between learnable LiDAR features and dense frozen vision CLIP features, using a single classification head to make predictions for both base and novel classes. To further improve the classification performance on novel classes and leverage the CLIP model, we propose two novel loss functions: object-level distillation loss and voxel-level distillation loss. Our experiments on the nuScenes and SemanticKITTI datasets show that our method outperforms the strong baseline by a large margin.
}

\end{abstract}
\vspace{-1mm}
\let\thefootnote\relax\footnotetext{$*$ Work done while at Waymo}    
\section{Introduction}
\label{sec:intro}
3D panoptic segmentation is a crucial task in computer vision with many real-world applications, most notably in autonomous driving. It combines 3D semantic and instance segmentation to produce per-point predictions for two different types of objects: \textit{things} (\eg, car) and \textit{stuff} (\eg, road). To date, there has been significant progress in 3D panoptic segmentation ~\cite{Zhou2021PanopticPolarNet,sirohi2021efficientlps, razani2021gp, li2022panoptic, xu2022sparse, xiao2023p3former}. Most recently, methods such as~\cite{xiao2023p3former} produce panoptic segmentation predictions directly from point clouds by leveraging learned queries to represent objects and Transformer-based~\cite{vaswani2017attention} architectures~\cite{caesar2020nuscenes,behley2019iccv} to perform the modeling. \begin{wrapfigure}{r}{0.59\textwidth}
 \vspace{-4.2mm} 
 \begin{center}
 \hspace{-2mm}
  \includegraphics[width=0.59\textwidth]{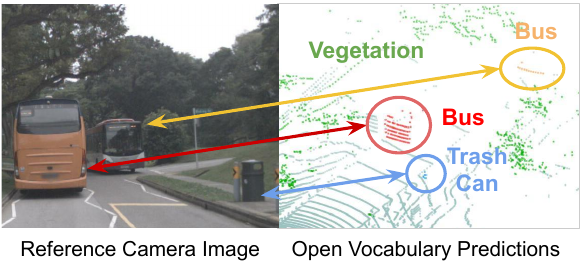}
 \end{center}
 \vspace{-6mm}
 \caption{ An illustration of 3D open-vocabulary panoptic segmentation results from our model. Without training on the categories of bus, trash can or vegetation, our method can produce accurate panoptic segmentation results even when the points are sparse.}
 \vspace{-7mm}
 \label{fig:teaser}
\end{wrapfigure}
 However, existing models only predict panoptic segmentation results for a closed-set of objects. They fail to create predictions for the majority of unseen object categories in the scene, hindering the application of these algorithms to real-world scenarios, 
especially for autonomous driving. In this work, we focus on segmenting unseen \textit{$things$} and unseen \textit{stuff} objects in autonomous driving scenarios.  Concretely, we follow~\cite{ding2023pla,yang2023regionplc} and develop models under the open-vocabulary setting: we divide the object categories into base (seen) categories and novel (unseen) categories, and evaluate models that are only trained on base categories. 
 
Such open-world computer vision tasks~\cite{bendale2015towards} benefit from the recent advancements in vision-language (V-L) models~\cite{radford2021learning,jia2021scaling}.
In 2D vision, there are many successful methods in open-vocabulary object detection~\cite{gu2021open, du2022learning, kuo2022f} and segmentation~\cite{ding2023open,xu2023open, yu2023fcclip}. These methods make predictions in a shared image-text embedding space, where predictions for unseen categories are produced by comparing the cosine similarity of an object with the text embedding of the category name. However, these methods are only possible due to the vast amounts of paired image-text data available, making it difficult to train similar models for 3D data such as LiDAR.

Instead, researchers have continued to leverage the effectiveness of these 2D vision-language models for 3D with the help of pixel-point correspondences by running inference on 2D images and then aligning with the 3D features. These methods have achieved promising results on open-vocabulary semantic
segmentation~\cite{ding2023pla,peng2023openscene,yang2023regionplc,zhang2023clip} and instance segmentation~\cite{ding2023pla,takmaz2023openmask3d,yang2023regionplc}, individually. However, there are no methods that address the problem of 3D open-vocabulary panoptic segmentation, \ie, addressing both open-vocabulary semantic segmentation and open-vocabulary instance segmentation at the same time. The challenge lies in how to handle segmentation for novel \textit{things} and \textit{stuff} objects simultaneously.

{
3D open-vocabulary panoptic segmentation is a challenging problem, due to both the significant domain gaps between the camera and LiDAR modalities and existing unsolved problems in open-vocabulary segmentation. Many existing open-vocabulary works rely on similarities between text embeddings of class names and pre-trained V-L features to obtain associations between predictions and classes~\cite{peng2023openscene,zhang2023clip,takmaz2023openmask3d}. However, while projecting 2D V-L features to 3D can account for a large part of the scene, there are often many points unaccounted for due to unmatched pixel/point distributions and differing fields of view between sensors. Some 3D open-vocabulary works~\cite{ding2023pla,yang2023regionplc} apply contrastive learning to obtain better association between language and points, but they require extra captioning models and do not address the difficulties  of detecting novel \textit{stuff} classes.
}

In this work, we aim to address these two issues with a novel architecture for 3D open-vocabulary panoptic segmentation. Building on existing 3D closed-set panoptic segmentation methods, we train a learned LiDAR feature encoder in parallel with a frozen, pre-trained camera CLIP model. By fusing the 3D LiDAR features with the 2D CLIP features, our model is able to learn rich features throughout the entire 3D sensing volume, even if there are no camera features in certain regions. In addition, we apply a pair of novel distillation losses that allow the 3D encoder to learn both object-level and voxel-level features which live inside the CLIP feature space. This provides a learned module in 3D space which can directly be compared with text embeddings. These losses also provide useful training supervision to unknown parts of the scene where there would otherwise be no loss gradient.

{
With the proposed model and loss functions, our method significantly outperforms the strong baseline on multiple datasets. 

Our contributions are summarized as follows:}
\begin{itemize}
    \vspace{-1.5mm}
    \item {We present the first approach for 3D open-vocabulary panoptic segmentation in autonomous driving.}
    \item {We propose two novel loss functions, object-level distillation loss and voxel-level distillation loss to help segment novel  \textit{things} and novel \textit{stuff} objects.}
    \item {We experimentally show that our proposed method significantly outperforms that strong baseline model on both nuScenes and SemanticKITTI datasets.}
    \vspace{-2mm}
\end{itemize}

\section{Related Work}
\label{sec:relatedwork}

{
This work is closely related to 3D panoptic segmentation, 2D open-vocabulary segmentation, and 3D open-vocabulary segmentation.}

{
\xhdr{3D panoptic segmentation.} The goal of 3D panoptic segmentation is to group 3D points according to their semantics and identities. This is a challenging task and relies on a good representation of the 3D data ~\cite{qi2017pointnet,qi2017pointnet++,wu2023pointconvformer,hu2021learning,alonso20203d,tang2020searching,xu2021rpvnet}. Most panoptic segmentation models have separate branches for instance segmentation and semantic segmentation~\cite{tang2020searching,hong2021lidar, Zhou2021PanopticPolarNet, li2022panoptic}. By following DETR~\cite{carion2020end}, the recently proposed P3Former~\cite{xiao2023p3former} uses learnable queries and a transformer architecture to obtain state-of-the-art performance on multiple panoptic segmentation benchmarks. Although those closed-set methods achieve incredible results, they cannot predict the labels and masks for novel classes.}
\begin{figure*}[!t]
\begin{center}
\includegraphics[width=0.95\linewidth]{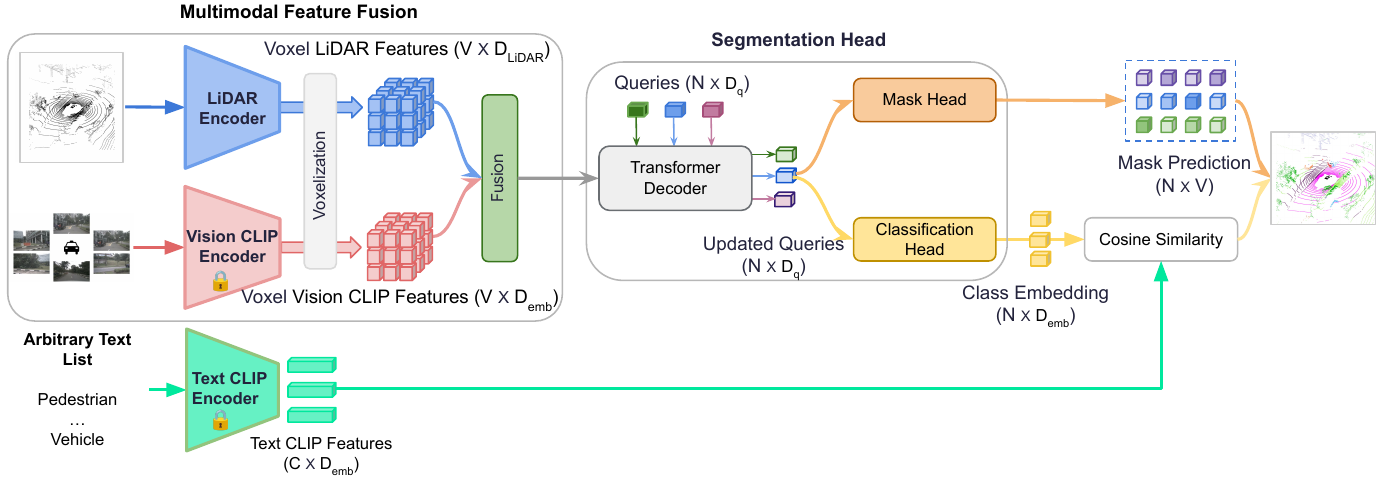}
\end{center}
\vspace{-15pt}
\caption{ Overview of our method. Given a LiDAR point cloud and the corresponding camera images, LiDAR features are extracted with a learnable LiDAR encoder, while vision features are extracted by a frozen CLIP vision model. The extracted LiDAR features and the frozen CLIP vision features are then fused and fed to a query-based transformer model to predict instance masks and semantic classes.
}
\label{fig:overall_fig}
\end{figure*}
{
\xhdr{2D open-vocabulary segmentation.} 2D open-vocabulary segmentation aims to group image pixels according to their semantics or identities for base (seen) or novel (unseen) categories. The prediction on novel categories is usually done by leveraging large V-L models~\cite{radford2021learning,jia2021scaling}. There are many works that focus on open vocabulary semantic segmentation~\cite{li2022languagedriven,xu2022groupvit, liu2022open,ghiasi2022scaling, xu2022simple, zhou2022extract, ma2022open, He_2023_CVPR,zhou2023zegclip,zou2023generalized,liang2023open}. Some work has also explored open-vocabulary panoptic segmentation~\cite{ding2023open,xu2023open,qin2023freeseg}. Recently, FC-CLIP~\cite{yu2023fcclip} proposes a single-stage framework based on a frozen convolutional CLIP backbone~\cite{liu2022convnet, radford2021learning,ig20215143773} for 2D open-vocabulary panoptic segmentation that achieves state-of-the-art performance. However, due to the camera-LiDAR domain gap, we show that simply extending it to 3D leads to poor performance.}

\xhdr{3D open-vocabulary segmentation.} 3D open-vocabulary segmentation is less explored due to the lack of 3D point-to-text association. One common practice is to utilize V-L models and use 2D-3D pairings to obtain rich, structured information in 3D~\cite{ha2022semantic,rozenberszki2022language,chen2023clip2scene,ding2023pla,yang2023regionplc,zhang2023clip,peng2023openscene,hegde2023clip,chen2023bridging,takmaz2023openmask3d}. Notably, CLIP2Scene\cite{chen2023clip2scene} proposes a semantic-driven cross-modal contrastive learning framework. PLA~\cite{ding2023pla} leverages images as a bridge and builds hierarchical 3D-caption pairs for contrastive learning. OpenScene~\cite{peng2023openscene} extracts per-pixel CLIP features using a pre-trained V-L model~\cite{li2022languagedriven, ghiasi2022scaling} then derives dense 3D features by projecting 3D points onto image planes. One concurrent work, RegionPLC~\cite{yang2023regionplc}, utilize regional visual prompts to create dense captions and perform point-discriminative contrastive learning, which is used for semantic segmentation or instance segmentation, individually. In contrast, our work does not rely on any captioning model or extra contrastive learning, but only depends on pre-trained CLIP features. Our model also simultaneously handles semantic and instance segmentation.

\section{Method}

\begin{figure*}[t]
\begin{center}
\includegraphics[width=\linewidth]{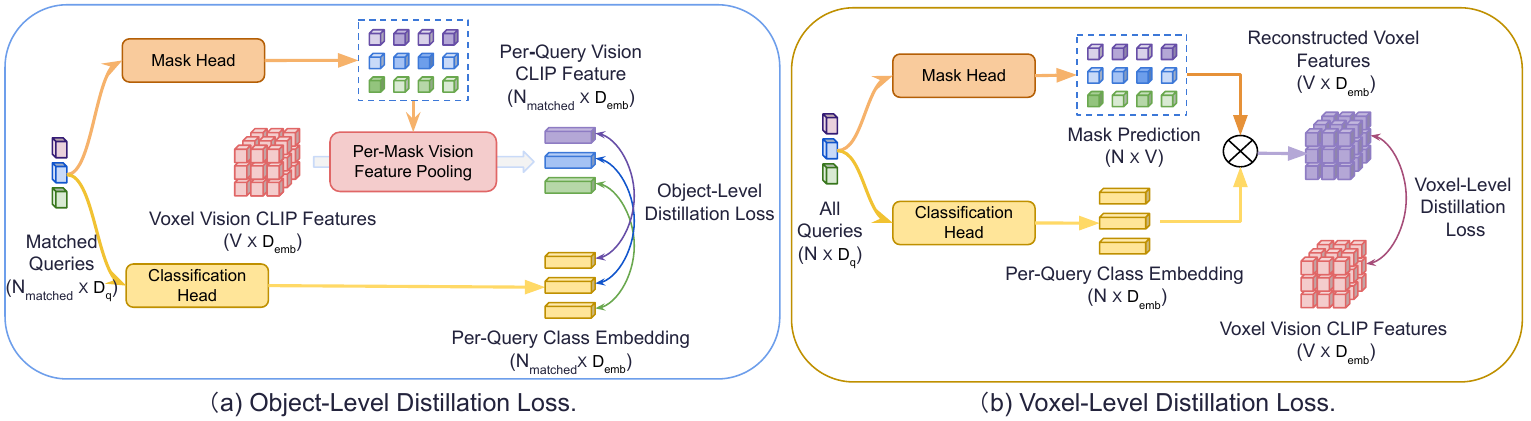}
\end{center}
\vspace{-15pt}
\caption{ (a) the proposed object-level distillation loss, and (b) the proposed voxel-level distillation loss.
}
\label{fig:loss_fig}
\end{figure*}

This section is organized as follows. First, we define the 3D open-vocabulary panoptic segmentation task. We then provide detailed descriptions of the model architecture as well as the proposed loss functions. The overview of our method is presented in~\cref{fig:overall_fig}, and the two proposed loss functions are illustrated in~\cref{fig:loss_fig} (a) and~\cref{fig:loss_fig} (b).

\subsection{Problem Definition}

{
In 3D panoptic segmentation, the goal is to annotate every point in a point cloud. For \textit{stuff} classes, (\eg road, vegetation), a category label is assigned according to its semantics. For \textit{things} classes (\eg cars, pedestrians), an instance label is assigned to objects in addition to its semantic label.}

In open-vocabulary panoptic segmentation, the models are trained on $C_B$ base(seen) categories. At test time, besides these $C_B$ base categories, the data will contain $C_N$ novel(unseen) categories. Following the settings of prior work~\cite{gu2021open,kuo2022f, yu2023fcclip}, we assume the availability of the name of the novel categories during inference, but the novel categories are not present in the training data and their names are not known. Note that we do not apply any prompt engineering, as this is not the focus of this paper. We follow OpenScene~\cite{peng2023openscene} to obtain the CLIP text embedding for each category.

\subsection{3D Open-Vocabulary Panoptic Segmentation}

{
{
Most of the previous 3D open-vocabulary works only address semantic segmentation~\cite{ha2022semantic,rozenberszki2022language,chen2023clip2scene,ding2023pla,yang2023regionplc,zhang2023clip,peng2023openscene,hegde2023clip,chen2023bridging} or instance segmentation~\cite{yang2023regionplc,takmaz2023openmask3d} separately, and there is no existing work for the 3D open-vocabulary panoptic segmentation task, which handles novel \textit{things} and novel \textit{stuff} 
objects simultaneously.} A natural idea would be extending the 2D open vocabulary segmentation methods to build the 3D counterpart. We start with P3Former~\cite{xiao2023p3former}, a state-of-the-art transformer-based 3D closed-set panoptic segmentation model, and add the essential components to support open-vocabulary capability by following FC-CLIP~\cite{yu2023fcclip}, a 2D open-vocabulary segmentation model that achieves state-of-the-art performance on  multiple datasets. However, we found that this simple extension leads to poor performance in our experiments, and in this work we propose several new features to improve the performance of our model. More implementation details for this baseline can be found in the supplementary material. 

In order to improve the open vocabulary capability of our model, we propose significant changes to the P3Former architecture, as well as two new loss functions. The architecture of our method is shown in~\cref{fig:overall_fig} and mainly consists of multimodal feature fusion, a segmentation head, and input text embeddings for open-vocabulary classification.
}

{
\xhdr{Multimodal feature fusion.} The core idea of many recent 2D open-vocabulary works is to leverage the features of large-scale vision-language models~\cite{radford2021learning,jia2021scaling}. These methods~\cite{yu2023fcclip} mainly rely on frozen CLIP features and use a transformer model to perform the 2D panoptic segmentation task. However, this is not optimal for 3D tasks since many points do not have corresponding valid camera pixels, leading to invalid features preventing meaningful predictions. To fully exploit the power of the CLIP vision features and learn complementary features from both CLIP features from camera and features from LiDAR, we generate predictions from the fusion of CLIP features extracted by a frozen CLIP model and learned LiDAR features from a LiDAR encoder.

As shown in~\cref{fig:overall_fig}, there are three major components for the multimodal feature fusion including a LiDAR encoder, a vision CLIP encoder, and voxel-level feature fusion. The LiDAR encoder is a model which takes an unordered set of points as input and extracts per-point features. We apply voxelization to the features from the LiDAR encoder, producing output features $F_{lidar} \in \mathbb{R}^{V\times D_{lidar}}$, where $V$ is the number of the voxels and $D_{lidar}$ is the dimension of the learned LiDAR feature. The Vision CLIP encoder is a pre-trained V-L segmentation model~\cite{ghiasi2022scaling} which extracts pixel-wise CLIP features from each camera image. Within each voxel, every LiDAR point is projected into the camera image plane based on the intrinsic and extrinsic calibration parameters to index into the corresponding vision CLIP features, then the vision CLIP features of all the points belonging to each voxel are averaged to represent that voxel. Zero padding is used for points which do not have any valid corresponding camera pixels. The voxel CLIP features will be referred as $F_{vclip} \in \mathbb{R}^{V\times D_{emb}}$, where $V$ is the number of voxels after voxelization and $D_{emb}$ is the dimension of the CLIP features. Finally, the learned per-voxel LiDAR features and frozen per-voxel vision CLIP features are concatenated together to be used as input into the transformer decoder in the segmentation head. This feature fusion enables our model to learn complementary information from both the LiDAR and CLIP features, allowing us to fine-tune our backbone for each dataset's specific data distribution.
}

{

\xhdr{Segmentation head.} The segmentation head is a transformer model that takes the LiDAR-Vision fused feature as input to produce panoptic segmentation results. Prior works, including existing 2D open-vocabulary works such as FC-CLIP~\cite{yu2023fcclip}, typically use learnable queries $q$ to represent each instance or thing, and they contain a mask prediction head $f_{mask}$ to produce the corresponding mask for each individual object and a classification head $f_{cls}$ to predict the per-mask class score for each known class. However, as a result, they also need to rely on another classifier to handle novel categories. Our goal is to use a single model to handle the prediction for both base and novel categories. Thus, we predict a class embedding instead of a class score for each mask. During training, the model learns to regress an analogy to the CLIP vision embedding for each mask, and the category prediction can be obtained by calculating its similarity with the CLIP text embedding of text queries during the inference stage. The class embedding $f_{cls}$ prediction is defined as: 
\begin{equation}
    v_{q} = f_{cls}(q) \in{\mathbb{R}^{D_{emb}}},
\label{eq:vq}    
\end{equation}
where $v_{q}$ is in the CLIP embedding space. The predicted class logits are then computed from the cosine similarity between the predicted class embedding and the text embedding of every category name from the evaluation set using a frozen CLIP model. The classification logits are defined as:
\begin{equation}
    s_{v_{q}} = \frac{1}{T}[\cos(v_{q},t_1),\cos(v_{q},t_2), \dots, \cos(v_{q},t_C)]
\label{eq:sq}
\end{equation}
where $t_i \in{\mathbb{R}^{D_{emb}}}$, $i\in\{1,2,\dots,C\}$ is the text embedding, $C$ is the number of categories ($C_B$ in training and $C_B$+$C_N$ in testing), and $T$ is a learnable temperature term that controls the concentration of the distribution.
}

\xhdr{Query assignment.} A common practice~\cite{yu2023fcclip, cheng2021per} for transformer-based panoptic segmentation models is to utilize a single set of queries to make predictions for both \textit{things} and \textit{stuff} classes jointly. In contrast, P3Former uses one query set to represent \textit{things} classes after bipartite matching and one fixed query set for \textit{stuff} classes. We have found that this separation of \textit{things} queries and \textit{stuff} queries makes our model converge faster and improve overall performance, and similar pattern has been observed in other tasks~\cite{Li_2022_CVPR}. However, the fixed set of queries for \textit{stuff} classes is not applicable to the open-vocabulary setting due to the unknown number of novel stuff classes. To take advantage of the benefits of separating \textit{things} queries and \textit{stuff} queries, we propose to predict the base \textit{stuff} classes with a fixed set of queries and utilize a set of learnable queries to target base \textit{things} classes and all novel (\textit{things} and \textit{stuff}) classes. More details of the query assignment can be found in the supplementary materials.

\subsection{Loss Function}

Closed-set panoptic segmentation models~\cite{xiao2023p3former} are typically optimized with objective functions consisting of a classification loss $L_{cls}$ and a mask prediction loss $L_{mask}$. We follow P3Former~\cite{xiao2023p3former} for these two losses: the classification loss $L_{cls}$ optimizes the focal loss~\cite{Lin2017Focal} between the class predictions and the category labels, while the mask loss $L_{mask}$ optimizes the voxel-query classification loss. Besides the two standard loss functions, we propose two simple yet effective losses to apply distillation from the CLIP model at different levels.

\xhdr{Object-level distillation loss.} Similar to previous methods~\cite{xu2023open,yu2023fcclip}, our method uses the cosine similarity between predicted class embeddings and class text CLIP embeddings to produce classification scores. However, the classification loss applied to ~\cref{eq:sq} only enforces similarity to known classes. In this work, we make the assumption that the frozen CLIP features are discriminative with respect to open-vocabulary classes and have good out-of-distribution generalization. We propose an additional training loss which forces our predicted object-level class embeddings to be similar to the CLIP embeddings within their corresponding masks after matching. Similar to~\cite{yu2023fcclip}, we utilize voxel vision CLIP features to get an  embedding for each query $q$ by mask pooling Vision CLIP features:
\begin{align}
    w_{q} =& \frac{1}{|M_q|}\sum_{p} \mathds{1}(p\in M_q) F_{vclip}(p)
\label{eq:wq}
\end{align}
where $M_q$ is the set of points $p$ belonging to the mask for query $q$. Our object-level distillation loss is then defined as:
\begin{equation}
    L_{O} = \frac{1}{|Q_{matched}|} \sum_{q\in Q_{matched}} 1 - \cos{(v_q,w_q)},
\end{equation}
where $Q_{matched}$ is the set of queries matched with ground truth objects during training, $v$ is the set of predicted class embeddings, and $w$ is the set of mask-pooled CLIP embeddings. This loss forces the model to directly distill object-level camera CLIP features and improves model performance for novel \textit{things} classes. We also experimented with applying $L_{O}$ to all predicted masks, but we found that this reduced model performance, likely due to the presence of masks that do not correspond to any objects in the scene.

\xhdr{Voxel-level distillation loss.}
{
{While the object-level distillation loss distills the per-object features from CLIP model, it does not provide any supervision for the mask prediction head, which would otherwise only receive supervision for known classes. We found this particularly problematic for unknown \textit{stuff} classes, which tend to be more spread out and cover larger and more diverse parts of the scene. In addition, it is only being applied to queries with relatively accurate mask predictions in order to learn useful CLIP features. To target these issues, we propose the voxel-level distillation loss to explicitly learn voxel-level CLIP features, which do not depend on any labels and can be applied on all queries.} In particular, the voxel-level distillation loss is defined as:
\begin{equation}
    F_{rec} = M_{Q}^{T}  F_{Qemb}
\end{equation}
where $Q$ is the number of queries, $F_{Qemb} \in \mathbb{R}^{Q\times{D_{emb}}}$ is the predicted embedding for all queries and $M_{Q} \in \mathbb{R}^{Q \times V}$ is the predicted per-voxel mask probabilities for all queries. The reconstructed features can be regarded as the weighted sum of all queries for each voxel. We then again supervise these features with the voxel CLIP features:
\begin{equation}
    L_{V} = L_1(F_{rec},F_{vclip})
\end{equation}
{Unlike the object-level distillation loss, which is only applied to queries with matched ground truth, this loss is applied to all predicted mask scores and queries. In our experiments, we found that this loss significantly improves performance on novel \textit{stuff} categories in particular, likely as it does not require exact matches with the ground truth, which can be difficult for large \textit{stuff} classes. However, this loss is still susceptible to noisy or low quality mask scores, and we found that larger weights for this loss can disrupt training.}

}

{
To summarize, $L_O$ helps get rid of the ensemble of classifiers in~\cite{gu2021open,ghiasi2022scaling,kuo2022f,xu2023open,yu2023fcclip} and enables open-vocabulary ability with one trainable classifier. $L_V$ uses a scene-level representation represented by the embedding of all queries, while previous methods only consider object-level representation. Combining $L_O$ with $L_V$ enables segmenting novel \textit{things} and novel \textit{stuff} objects simultaneously, which makes us the first work to address the 3D open-vocabulary panoptic segmentation problem in autonomous driving.
} Our final objective function can be written as:
\begin{equation}
    L = w_\alpha * L_{cls}+ w_\beta * L_{mask}+ w_\lambda * L_{O}+ w_\gamma * L_{V}
\end{equation}, where $w_\alpha$, $w_\beta$, $w_\lambda$, $w_\gamma$, are weights for the corresponding objective function.

\subsection{Implementation Details}

For the LiDAR encoder and segmentation head, we follow the implementation of the state-of-the-art closed-set 3D panoptic segmentation method P3Former. For the Vision CLIP encoder, we use OpenSeg\cite{ghiasi2022scaling}, due to its remarkable performance on the recent open-vocabulary 3D semantic segmentation task~\cite{peng2023openscene}. For the Text CLIP encoder, we use CLIP\cite{radford2021learning} with ViT-L/14\cite{vaswani2017attention} backbone, following other state-of-the-art open vocabulary works~\cite{peng2023openscene}.

\section{Experiments}
\label{sec:experiments}

{
Following the state-of-the-art closed-set 3D panoptic segmentation work~\cite{Zhou2021PanopticPolarNet,sirohi2021efficientlps, razani2021gp, li2022panoptic, xu2022sparse, xiao2023p3former}, we conduct experiments and ablation studies on the nuScenes~\cite{caesar2020nuscenes} and SemanticKITTI~\cite{geiger2012cvpr,behley2019iccv} datasets.}

\subsection{Experimental Setting}

{
\xhdr{nuScenes.} The nuScenes dataset~\cite{caesar2020nuscenes} is a public benchmark for autonomous driving. It consists of $1000$ run segments and is further divided into prescribed train/val/test splits. We use all key frames with panoptic labels in the training set($28130$ frames) to train the model. Following the most recent state-of-the-art model P3Former~\cite{xiao2023p3former}, we evaluate the models on the validation set($6019$ frames). There are $16$ semantic classes, including $10$ \textit{things} classes and $6$ \textit{stuff} classes.}

{
\xhdr{SemanticKITTI.} SemanticKITTI~\cite{geiger2012cvpr,behley2019iccv} is the first large dataset for LiDAR panoptic segmentation for autonomous driving. We conduct experiments on the training and validation sets, where panoptic segmentation labels are available. 3D open-vocabulary methods often require point and pixel pairing. In the SemanticKITTI dataset, however, the ego-vehicle is only equipped with frontal cameras. Thus, we filter out the points that are not visible in the camera view based on the provided camera parameters for both training and evaluation. There are $19$ semantic classes, including $8$ \textit{things} classes and $11$ \textit{stuff} classes.
}

{
\xhdr{Data split.} Both the nuScenes and SemanticKITTI datasets do not provide official base and novel class splits. Following the state-of-the-art 3D open-vocabulary segmentation work~\cite{cen2022openworld,ding2023pla,yang2023regionplc}, we randomly split the classes into base and novel, while keeping the ratio between base and novel classes around $3:1$. For nuScenes, the number of class for base and novel split are $12$ and $4$ respectively, and this setting will be referred as B12/N4. For SemanticKITTI, the number of class for base and novel split are $14$ and $5$, and this setting will be referred as B14/N5. We use the same splits in the main comparison with prior methods, and provide the results of more variations in the ablation studies and supplementary materials.
}

{
\xhdr{Training details.} We follow most of the architecture configurations in the official P3Former\cite{xiao2023p3former} implementation. We set $w_\alpha=1$, $w_\beta=1$, $w_\lambda=1$, $w_\gamma=0.1$ for both datasets. We use the AdamW~\cite{kingma2014adam,loshchilov2017decoupled} optimizer with a weight decay of $0.01$. We set the initial learning rate as $0.0008$ with a multi-step decay schedule. The models are trained for $40$ epochs, and we use the checkpoint of the last epoch for evaluation. To avoid ambiguous class names and better utilize the CLIP text embedding, we follow~\cite{lambert2020mseg,peng2023openscene,yu2023fcclip} and apply multi-label mapping for the text queries. During inference, if there are multiple labels for one class, we derive the class score by getting the maximum scores among these labels.
}

{
\xhdr{Evaluation metrics.} We use panoptic quality ($PQ$) as the major evaluation metric for the panoptic segmentation task. $PQ$ is formulated as:
\eqnsm{psq-seg-det}{\small{\text{PQ}} = \underbrace{\frac{\sum_{\TP} \text{IoU}}
{|\TP|}}_{\text{SQ}} \times \underbrace{\frac{|\TP|}{|\TP| + \frac{1}{2} |\FP| + \frac{1}{2} |\FN|}}_{\text{RQ}}.
\vspace{-2mm}
}
$PQ$ is the multiplication of segmentation quality ($SQ$) and recognition quality ($RQ$). We report all the three metrics ($PQ$, $RQ$, $SQ$) for all classes.} We also report $PQ$, $RQ$, $SQ$ for novel \textit{things} objects and novel \textit{stuff} objects separately. In particular, $PQ_{N}^{Th}$ means $PQ$ for novel \textit{things} classes and $PQ_{N}^{St}$ stands for $PQ$ for novel \textit{stuff} classes. We also report the mean Intersection over Union (mIoU) for all classes to measure semantic segmentation quality.

\subsection{P3Former-FC-CLIP Baseline}
As a baseline for novel-class panoptic segmentation, we construct a model from a fusion of P3Former~\cite{xiao2023p3former} and FC-CLIP~\cite{yu2023fcclip}. This baseline will be called P3Former-FC-CLIP (PFC). The baseline model takes the frozen voxel vision CLIP features as input, and the final prediction is obtained by geometric ensembling\cite{gu2021open,ghiasi2022scaling,kuo2022f,xu2023open,yu2023fcclip} of the results from the classification head $f_{cls}$ and another frozen classifier based on the similarity between the average-pool class embedding $w_q$ and the CLIP text embedding. Following FC-CLIP~\cite{yu2023fcclip}, the same set of learnable queries were used to represent both \textit{things} and \textit{stuff} classes. 
%
In summary, this baseline provides a comparison against our proposed method without the multimodal feature fusion module, the unified segmentation head, and the distillation losses. More information on this baseline can be found in the supplementary material.

\begin{figure}[t]
\begin{center}
\includegraphics[width=\linewidth]{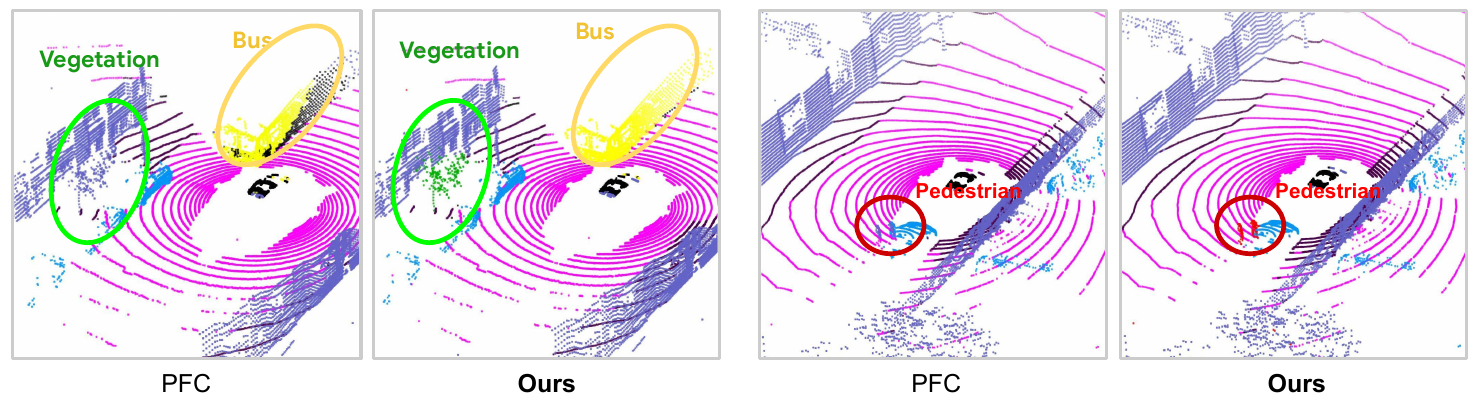}
\end{center}
\vspace{-20pt}
\caption{ Open-vocabulary panoptic segmentation results from PFC and our method on nuScenes. PFC predicts inaccurate category and masks for the novel pedestrian (red), bus (yellow) and vegetation (green), while ours makes correct predictions.
}

\label{fig:visualization}
\end{figure}
\subsection{Main Results}

{
Since there are no existing methods for the 3D open-vocabulary panoptic segmentation task, we mainly compare with three methods to demonstrate the capability of our method: (1) the strong open-vocabulary baseline method PFC to fairly demonstrate the strength of our method, (2) the closed-set state-of-the-art 3D panoptic segmentation method P3Former to understand the headroom of our method, and (3) the open-set, zero-shot state-of-the-art method for 3D semantic segmentation, OpenScene~\cite{peng2023openscene}. Comparisons on the nuScenes and SemanticKITTI datasets are shown in ~\cref{table:ns} and ~\cref{table:kitti}.
}

{

{

\xhdr{Results on nuScenes dataset.} \cref{table:ns} shows the performance comparison on the validation set of the nuScenes dataset.  Our method significantly outperforms the strong baseline PFC by a large margin across all metrics. PFC works relatively well for the novel \textit{things} classes, but performance on the novel \textit{stuff} class collapses. This is likely because \textit{stuff} classes tend to cover large parts of the scene, leading to diverse per-voxel CLIP features which may not be good representatives for their respective classes. For the novel \textit{stuff} classes, PFC only achieves $0.5$ $PQ_{N}^{St}$, while our method achieves $35.2$. PFC can produce reasonable masks for the novel \textit{stuff} classes (with $SQ_{N}^{St}$=$60.4$), but fails to produce high quality semantic predictions for the novel \textit{stuff} class (with $RQ_{N}^{St}$= $0.8$).  \cref{fig:visualization} shows a qualitative comparison between PFC and our method. With our proposed design, and especially the new loss functions, our method is able to achieve strong performance for both novel \textit{things} and \textit{stuff} classes.}
\begin{table*}[t]
\small
\centering
\caption{ \small \textbf{Quantitative results of panoptic segmentation on nuScenes}. We compare the performance of open-vocabulary and fully supervised models. All open vocabulary models share the same randomly picked base/novel split: B12/N4. The novel \textit{things} classes are bus, pedestrian and motorcycle. The novel \textit{stuff} class is vegetation.}
\vspace{-5mm}
\renewcommand{\arraystretch}{1.0}
	\vspace{2mm}
	\newcommand{\tabincell}[2]{\begin{tabular}{@{}#1@{}}#2\end{tabular}}
	\centering
	\renewcommand{\arraystretch}{1.43}
	\resizebox{\textwidth}{!}
	{
		\centering
            \begin{tabular}{c|c|c|c|c|c|c|c|c|c|c|c|c}
            \specialrule{.2em}{.1em}{.1em}
            Model & Type & Supervision  & $PQ$  & $ PQ_{N}^{Th} $ & $ PQ_{N}^{St} $  & $RQ$ & $ RQ_{N}^{Th} $ & $ RQ_{N}^{St} $ & $SQ$  & $ SQ_{N}^{Th} $ & $ SQ_{N}^{St} $ & mIoU\\ 
            \hline
            P3Former~\cite{xiao2023p3former} &  closed-set  & full    & 75.9   & 85.1 & 82.9 & 84.7  & 89.9  & 95.9 & 89.8   & 94.7 & 86.5  &   76.8\\ 
            \hline
            \hline
            OpenScene~\cite{peng2023openscene} &   open-voc  & zero-shot    & -   & - & - & -  & -  & - & -   & - & - &  42.1 \\ \hline
            PFC &  open-voc  & partial 	& 54.8	& 37.3	& 0.5	& 63.6	& 42.1	& 0.8	& 84.2 & 	\textbf{89.3} & 60.4 & 55.5\\ \hline
            Ours &  open-voc  & partial                   & \textbf{62.0}                    & \textbf{49.6}                  & \textbf{35.2}                  & \textbf{70.9}                  & \textbf{55.6}                 & \textbf{46.0}                  & \textbf{87.0}                    & 89.1                  & \textbf{76.7}  &  \textbf{60.1} \\ 
            \hline
            \end{tabular}
        }
\vspace{-1mm}
\label{table:ns}
\end{table*}

\begin{table}[t]

\tabcolsep=0.13cm
\small
\centering
\caption{
\textbf{Performance for base classes on nuScenes}. We report the performance on base classes for models in~\cref{table:ns}. A gap still exists between open and closed-set methods for base classes. We show that this is due to lack of supervision of the whole scene as P3Former achieves similar performance when only trained on base categories.}
\renewcommand{\arraystretch}{1.0}
	\vspace{-3mm}
	\newcommand{\tabincell}[2]{\begin{tabular}{@{}#1@{}}#2\end{tabular}}
	\centering
	\renewcommand{\arraystretch}{1.2}
	\resizebox{0.9\textwidth}{!}
	{

    \begin{tabular}{c|c|c|ccc|ccc}
    \specialrule{.2em}{.1em}{.1em}
    \multirow{2}{*}{Model} & \multirow{2}{*}{Supervision} & \multirow{2}{*}{Training Data} & \multicolumn{3}{c|}{Base 
    \textit{Things}}                                & \multicolumn{3}{c}{Base \textit{Stuff}}                                 \\ \cline{4-9} & &
                            & \multicolumn{1}{c|}{$PQ_B^{Th}$}    & \multicolumn{1}{c|}{$RQ_B^{Th}$}    & $SQ_B^{Th}$    & \multicolumn{1}{c|}{$PQ_B^{St}$}    & \multicolumn{1}{c|}{$RQ_B^{St}$}    & $SQ_B^{St}$    \\ \hline
    P3Former~\cite{xiao2023p3former} & full   & base+novel            & \multicolumn{1}{c|}{73.4} & \multicolumn{1}{c|}{80.5} & 90.9 & \multicolumn{1}{c|}{73.9}  & \multicolumn{1}{c|}{85.3} & 85.9 \\ 
    \hline

    P3Former~\cite{xiao2023p3former}  & partial   & base            & \multicolumn{1}{c|}{65.2} & \multicolumn{1}{c|}{71.3} & 88.0 & \multicolumn{1}{c|}{64.2}  & \multicolumn{1}{c|}{77.4} & 81.8 \\ 
    \hline
    \hline
    PFC   & partial     & base       & \multicolumn{1}{c|}{65.6} & \multicolumn{1}{c|}{73.3} & 89.0 & \multicolumn{1}{c|}{61.0} & \multicolumn{1}{c|}{75.4} & 83.7 \\ \hline
    Ours   & partial     & base               & \multicolumn{1}{c|}{66.7} & \multicolumn{1}{c|}{73.7} & 89.8 & \multicolumn{1}{c|}{69.2} & \multicolumn{1}{c|}{82.1} & 83.7 \\ \hline
    \end{tabular}
}
\label{table:base}
\end{table}

{
To further understand the headroom of our method, we also compare our model with the closed-set P3Former. Note that the comparison here is deliberately unfair since the supervision signals are different.
Compared with the closed-set P3Former, our segmentation quality($SQ$) is good while there is a large gap on mask classification quality($RQ$).
 
The gap is largely due to regressions in the novel classes, where precise supervision is not available for open-vocabulary models. For base classes, as shown in~\cref{table:base}, the gap is relatively small except for a drop in $RQ_B^{Th}$. We believe the closed-set P3Former sees ground truth supervision for the entire scene, while the open-set methods do not receive supervision in the ‘unknown class’ regions. In fact, when P3Former is only trained on base categories, the performance is worse than our proposed method.
}

Besides the comparison with the closed-set method, we also compare with the zero-shot state-of-the-art method OpenScene~\cite{peng2023openscene} which does not use any labels for training. In this comparison, our model significantly outperforms OpenScene in the mIoU metric for semantic segmentation. Note that this comparison is not entirely fair, as our method is trained with partial labels. Instead, the comparison is useful to understand the gap between the two types of open-vocabulary methods. {
The concurrent work RegionPLC~\cite{yang2023regionplc} also reports open-vocabulary results for the semantic segmentation task on the nuScenes dataset. However, we cannot directly compare with this method since it removes one class (other-flat)
and does not provide its base/novel split.
}

}

{
\begin{table*}[t]
\small
\centering
\caption{\textbf{Quantitative results of panoptic segmentation on SemanticKITTI}. We compare the performance of open-vocabulary and fully supervised models. All open vocabulary models share the same randomly picked base/novel split: B14/N5. The novel \textit{things} classes are bicycle and truck. The novel \textit{stuff} classes are sidewalk, building and trunk.}
\renewcommand{\arraystretch}{1.0}
	\vspace{-3mm}
	\newcommand{\tabincell}[2]{\begin{tabular}{@{}#1@{}}#2\end{tabular}}
	\centering
	\renewcommand{\arraystretch}{1.43}
	\resizebox{\textwidth}{!}
	{
        \begin{tabular}{c|c|c|c|c|c|c|c|c|c|c|c|c}
        \specialrule{.2em}{.1em}{.1em}
        Model & Type & Supervision  & $PQ$  & $ PQ_{N}^{Th} $ & $ PQ_{N}^{St} $  & $RQ$ & $ RQ_{N}^{Th} $ & $ RQ_{N}^{St} $ & $SQ$  & $ SQ_{N}^{Th} $ & $ SQ_{N}^{St} $ & mIoU\\ 
        \hline
        P3Former~\cite{xiao2023p3former} &  closed-set  & full     & 62.1   & 65.9 & 74.2 & 71.3  & 74.8  & 86.8 & 77.1   & 88.3 & 83.9 &  61.6 \\
        \hline
        \hline
        PFC &  open-voc  & partial 	& 33.7	& 12.0	& 0.4	& 40.1	& 15.0	& 0.6	& 67.6 & 81.1 & 47.3 & 33.4 \\ 
        \hline
        Ours &  open-voc  & partial                   & \textbf{42.2}                    & \textbf{13.1}                  & \textbf{17.8}                  & \textbf{50.4}                  & \textbf{16.2}                 & \textbf{26.7}                  & \textbf{73.0}                    & \textbf{84.0}                  & \textbf{67.2}  &  \textbf{44.6} \\ 
        \hline
        \end{tabular}
        
    }

\label{table:kitti}
\end{table*}
\xhdr{Results on SemanticKITTI dataset.} To demonstrate the generalization ability of our method across different datasets, we report the results on SemanticKITTI dataset in~\cref{table:kitti}. Overall, we observe similar patterns as on the nuScenes dataset. The baseline achieves relatively poor overall performance and struggles with the novel \textit{stuff} classes. Using our architecture and loss functions, our model significantly outperforms PFC on $PQ$, with the largest margin for novel \textit{stuff} classes. Note that the gap between the open-vocabulary methods (ours and PFC) and the closed-set method is larger on SemanticKITTI, likely due to the smaller dataset limiting performance.
}

\subsection{Ablation Studies and Analysis}
\begin{table*}[t]
\small
\centering
\renewcommand{\arraystretch}{1.0}
	\newcommand{\tabincell}[2]{\begin{tabular}{@{}#1@{}}#2\end{tabular}}
	\centering
    \caption{\textbf{Impact of each component}. We evaluate the impact of each component using the base/novel split in \cref{table:ns}. We observe that each component can provide improvements over the PCF baseline. Noticeably, $L_{V}$ brings the biggest improvement.}
    \vspace{-3mm}
	\renewcommand{\arraystretch}{1.2}
	\resizebox{\textwidth}{!}
	{
		\centering
            \begin{tabular}{ccllc|c|c|c|c|c|c|c|c|c}
            \specialrule{.2em}{.1em}{.1em}
            \multicolumn{4}{c|}{Components}                                                                                                             & \multirow{2}{*}{~$PQ$~}   & \multirow{2}{*}{ ~$PQ_{N}^{Th} $~} & \multirow{2}{*}{~$PQ_{N}^{St~} $} & \multirow{2}{*}{~$RQ$~}   & \multirow{2}{*}{~$RQ_{N}^{Th} $~} & \multirow{2}{*}{~$RQ_{N}^{St} $~} & \multirow{2}{*}{~$SQ$~}   & \multirow{2}{*}{~$SQ_{N}^{Th} $~} & \multirow{2}{*}{~$SQ_{N}^{St} $~}  &           \multirow{2}{*}{~mIoU~} \\ \cline{1-4}
            \multicolumn{1}{c|}{QA} & \multicolumn{1}{c|}{Fusion} & \multicolumn{1}{c|}{$L_{O}$} & \multicolumn{1}{c|}{$L_{V}$} &                   &                       &                       &                       &                       &                       &                       &                       &                       &                       \\ \hline
            \multicolumn{1}{l|}{}                 & \multicolumn{1}{l|}{}       & \multicolumn{1}{l|}{}       & \multicolumn{1}{l|}{}       	& 54.8	& 37.3	& 0.5	& 63.6	& 42.1	& 0.8	& 84.2 & 	\textbf{89.3} & 60.4  & 55.5                  \\ \hline
            \multicolumn{1}{c|}{\checkmark}                & \multicolumn{1}{l|}{}       & \multicolumn{1}{l|}{}       & \multicolumn{1}{l|}{}                        & 55.5                 & 35.7                 & 0.4                 & 64.0                 & 40.8                & 0.7                  & 84.3                 & 87.4                 & 56.5      & 56.6           \\ \hline
            \multicolumn{1}{c|}{\checkmark}                & \multicolumn{1}{c|}{\checkmark}      & \multicolumn{1}{l|}{}       & \multicolumn{1}{l|}{}                       & 56.4                 & 38.1                & 0.4                  & 65.0                 & 43.5                & 0.6                  & 84.6                 & 87.4                 & 61.3     & 56.4             \\ \hline
            \multicolumn{1}{c|}{\checkmark}                & \multicolumn{1}{c|}{\checkmark}      &  \multicolumn{1}{c|}{\checkmark}      & \multicolumn{1}{c|}{ }                        & 56.3                    & 43.8                  & 0.2                  & 64.8                  & 49.2                 & 0.3                    & 85.1                    & 88.9                  & 64.0             & 54.0     \\ \hline
            \multicolumn{1}{c|}{\checkmark}                & \multicolumn{1}{c|}{\checkmark}      &  \multicolumn{1}{c|}{\checkmark}      & \multicolumn{1}{c|}{\checkmark}                        & \textbf{62.0}                    & \textbf{49.6}                  & \textbf{35.2}                  & \textbf{70.9}                  & \textbf{55.6}                 & \textbf{46.0}                    & \textbf{87.0}                    & 89.1                  & \textbf{76.7}             & \textbf{60.1}     \\    \hline
        \end{tabular}
        }
        \vspace{2mm}
\vspace{-5mm}
\label{table:ablation}
\end{table*}
{
To better understand the effectiveness of each component, we conduct ablation studies for each design choice and loss function on the nuScenes dataset. These results are shown in~\cref{table:ablation}. We conduct five sets of experiments, starting with the PFC baseline 
and build upon it four ablations with different combinations.
}

{
\xhdr{Impact of query assignment.} Starting from the PFC baseline model, we add our proposed fixed query assignment for \textit{stuff} categories. The results for this ablation are shown in the second row of~\cref{table:ablation}. With query assignment, the overall $PQ$ improves by $0.7$, showing that it indeed helps the overall performance. The performance for the novel classes drop slightly, but improvement on the base classes overcomes this for the overall PQ.
}

{
\xhdr{Impact of feature fusion.} The third row of~\cref{table:ablation} shows the impact of feature fusion. Without feature fusion, our model already achieves $55.5$ $PQ$, demonstrating the power of the CLIP vision features. The third row shows that the performance with feature fusion for the model input improves the overall $PQ$ by $0.9$. This slightly improved the overall performance, but the improvement on the novel \textit{things} class is the most significant, demonstrating that the learned LiDAR features and CLIP vision features are indeed complementary for the task.
}

{
\xhdr{Impact of object-level distillation loss.} The fourth row of the results in~\cref{table:ablation} shows the impact of the proposed object-level distillation loss. Note that for models with the object-level distillation loss, we also remove the frozen class classification head and geometric ensemble in the PFC baseline, consolidating to a single class embedding head. Although the $RQ_{N}^{St}$ slightly dips by $0.3$ for the novel \textit{stuff} classes, this loss can significantly improve the $RQ_{N}^{Th}$ for the novel \textit{things} class by $5.7$. Moreover, this loss helps improve the segmentation quality ($SQ$) for both \textit{things} and \textit{stuff} classes.
}

{
\xhdr{Impact of voxel-level distillation loss.} We further study the impact of the voxel-level distillation loss to see if it can further improve the performance given all of our designs. The results are shown in the last row of ~\cref{table:ablation}. With this loss function, $PQ$ significantly improves by $5.7$. The improvement on the novel split is particularly large, especially for the novel \textit{stuff} classes. The $PQ_{N}^{St}$ of the novel \textit{stuff} class improves from $0.2$ to $35.2$ demonstrating the importance of the voxel-level supervision to the performance of the novel \textit{stuff} class.
}
\begin{table*}[t]

\small
\centering
\caption{\textbf{Performance of panoptic segmentation on nuScenes with a different split}. We compare the performance with a different split with 5 novel classes (B11/N5). The novel \textit{things} classes are bicycle, car and construction vehicle. The novel \textit{stuff} classes are terrain and man-made. Our method consistently outperforms the PFC baseline across all the metrics by a large margin.}
\renewcommand{\arraystretch}{1.0}
	\vspace{-3mm}
	\newcommand{\tabincell}[2]{\begin{tabular}{@{}#1@{}}#2\end{tabular}}
	\centering
	\renewcommand{\arraystretch}{1.43}
	\resizebox{\textwidth}{!}
{
    \begin{tabular}{c|c|c|c|c|c|c|c|c|c|c|c|c}
    \specialrule{.2em}{.1em}{.1em}
    Model & Type & Supervision  & $PQ$  & $ PQ_{N}^{Th} $ & $ PQ_{N}^{St} $  & $RQ$ & $ RQ_{N}^{Th} $ & $ RQ_{N}^{St} $ & $SQ$  & $ SQ_{N}^{Th} $ & $ SQ_{N}^{St} $ & mIoU\\ 
    \hline
    P3Former\cite{xiao2023p3former} &  closed-set  & full     & 75.8   & 70.5 & 71.7 & 83.8  & 76.4  & 85.5 & 90.1   & 91.6 & 83.6 &  75.0\\ 
    \hline
    \hline
    PFC &  open-voc  & partial 	& 43.9	& 27.7	& 0.6	& 51.7	& 33.2	& 1.0	& 80.2 & 82.4 & 62.7 & 45.2 \\ \hline
    Ours &  open-voc  & partial                   & \textbf{52.8}                    & \textbf{56.0}                  & \textbf{16.4}                  & \textbf{60.5}                  & \textbf{61.8}                 & \textbf{22.6}                  & \textbf{84.9}                    & \textbf{89.7}                 & \textbf{68.7}  &  \textbf{49.9} \\ 
    \hline
    \end{tabular}
}
\label{table:ns_s2}
\end{table*}
\begin{figure}[t]
\begin{center}
\includegraphics[width=\linewidth]{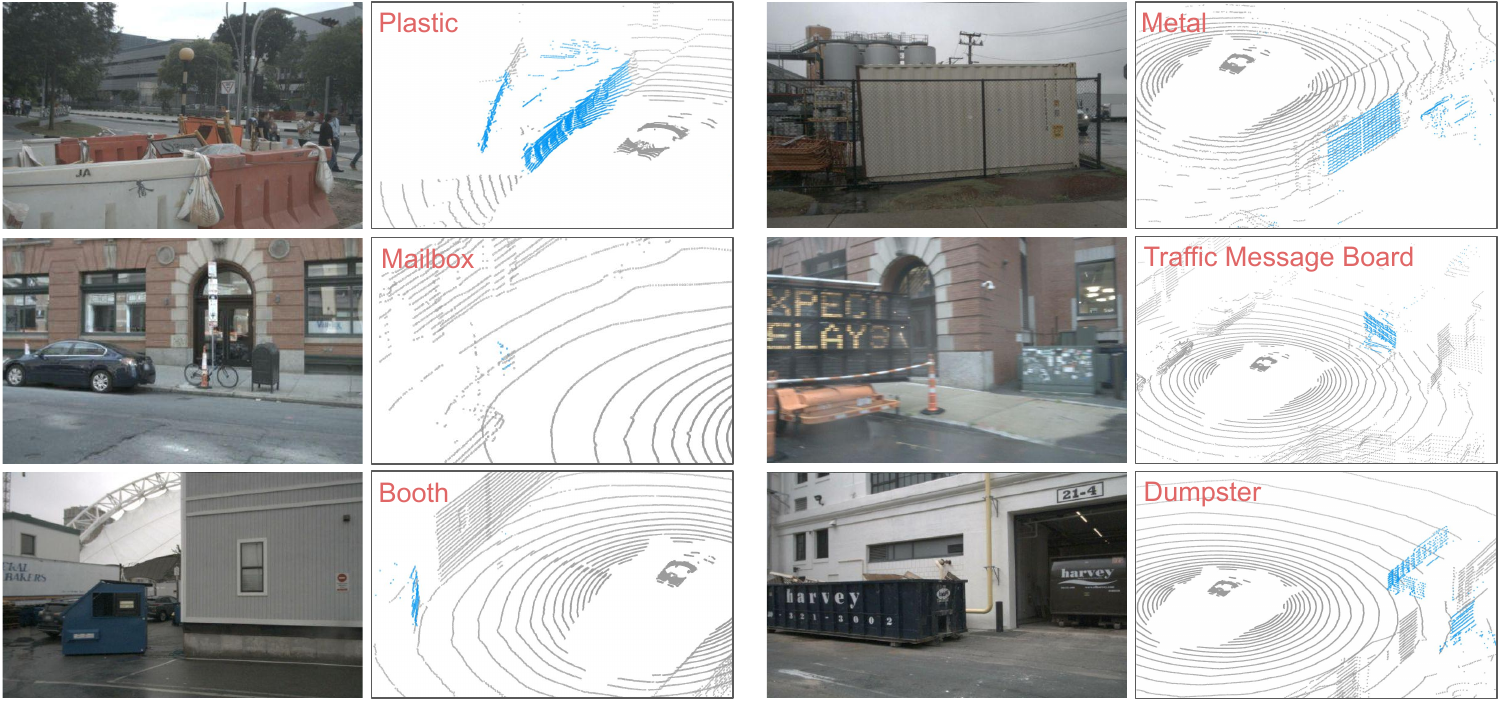}
\end{center}
\caption{
Open-vocabulary exploration on nuScenes. We show the novel material/object in blue color. The orientation of the ego vehicle is fixed in the LiDAR point visualization while the reference images come from on of the surrounding cameras of the ego vehicle.}
\label{fig:property}

\end{figure}

{
\xhdr{Performance of different splits.}
To validate the generalizability of our method across different base/novel class split, we conduct experiments on a different split (B11/N5) for the nuScenes dataset. Here, the five novel categories are bicycles, cars, construction vehicles, terrain, and manmade. As shown in~\cref{table:ns_s2}, similar to the results on the other split (B12/N4), our proposed method consistently and significantly outperforms the strong baseline method by a large margin. This again demonstrates the effectiveness of our design and the proposed loss functions. However, we do notice that the performance of our method and the baseline drops when more classes are included into the novel split.}

{

\xhdr{Open-vocabulary exploration.} In previous experiments, we follow other 3D open-vocabulary works~\cite{cen2022openworld,ding2023pla,yang2023regionplc} and provide analytical results on pre-defined object categories, mainly due to the limited categories in current panoptic segmentation datasets. In practice, our model goes beyond detecting these object categories: we can take class embeddings $v_q$ in~\cref{eq:vq} and compute the cosine similarity with CLIP embedding of any text.~\cref{fig:property} shows that we can detect novel materials/objects that are not in the predefined category list.
Note that the concept of open vocabulary is very different from domain adaptation, as open vocabulary refers to the ability to deal with novel inputs in a scene while domain adaptation addresses the difference in data distributions in different scenes.

\xhdr{Limitations.} Our models are only evaluated on current autonomous driving panoptic segmentation benchmarks, where a limited number of category annotations are available. To further quantitatively evaluate open-vocabulary performance, a large-scale autonomous driving benchmark with more diverse object categories is greatly desired.
}
\section{Conclusion}

{
In this paper, we present the first approach for the open vocabulary 3D panoptic segmentation task in autonomous driving by leveraging large vision-language models. We experimentally verified that simply extending the 2D open-vocabulary segmentation method into 3D does not yield good performance, and demonstrated that our proposed model design and loss functions significantly boost performance for this task. Our method significantly outperformed the strong baseline on multiple well-established benchmarks in the literature. We hope our work can shed light on the future studies of the 3D open-vocabulary panoptic segmentation problem.
}
\\
\\
\noindent\textbf{Acknowledgements.}{ We would like to thank Mahyar Najibi, Chao Jia, Zhenyao Zhu, Yolanda Wang, Charles R. Qi, Dragomir Anguelov, Tom Ouyang, Ruichi Yu, Chris Sweeney, Colin Graber, Yingwei Li, Sangjin Lee, Weilong Yang, and Congcong Li for the help to the project.}

\bibliographystyle{splncs04}
\bibliography{main}

\begin{thebibliography}{10}
\providecommand{\url}[1]{\texttt{#1}}
\providecommand{\urlprefix}{URL }
\providecommand{\doi}[1]{https://doi.org/#1}

\bibitem{alonso20203d}
Alonso, I., Riazuelo, L., Montesano, L., Murillo, A.C.: 3d-mininet: Learning a
  2d representation from point clouds for fast and efficient 3d lidar semantic
  segmentation. IEEE Robotics and Automation Letters  \textbf{5}(4),
  5432--5439 (2020)

\bibitem{behley2019iccv}
Behley, J., Garbade, M., Milioto, A., Quenzel, J., Behnke, S., Stachniss, C.,
  Gall, J.: {SemanticKITTI: A Dataset for Semantic Scene Understanding of LiDAR
  Sequences}. In: ICCV (2019)

\bibitem{bendale2015towards}
Bendale, A., Boult, T.: Towards open world recognition. In: CVPR (2015)

\bibitem{caesar2020nuscenes}
Caesar, H., Bankiti, V., Lang, A.H., Vora, S., Liong, V.E., Xu, Q., Krishnan,
  A., Pan, Y., Baldan, G., Beijbom, O.: nuscenes: A multimodal dataset for
  autonomous driving. In: CVPR (2020)

\bibitem{carion2020end}
Carion, N., Massa, F., Synnaeve, G., Usunier, N., Kirillov, A., Zagoruyko, S.:
  End-to-end object detection with transformers. In: ECCV (2020)

\bibitem{cen2022openworld}
Cen, J., Yun, P., Zhang, S., Cai, J., Luan, D., Wang, M.Y., Liu, M., Tang, M.:
  Open-world semantic segmentation for {LIDAR} point clouds. In: ECCV (2022)

\bibitem{chen2023clip2scene}
Chen, R., Liu, Y., Kong, L., Zhu, X., Ma, Y., Li, Y., Hou, Y., Qiao, Y., Wang,
  W.: Clip2scene: Towards label-efficient 3d scene understanding by clip. In:
  CVPR (2023)

\bibitem{chen2023bridging}
Chen, Z., Li, B.: Bridging the domain gap: Self-supervised 3d scene
  understanding with foundation models. arXiv preprint arXiv:2305.08776  (2023)

\bibitem{cheng2021per}
Cheng, B., Schwing, A., Kirillov, A.: Per-pixel classification is not all you
  need for semantic segmentation. In: NeurIPS (2021)

\bibitem{ding2023pla}
Ding, R., Yang, J., Xue, C., Zhang, W., Bai, S., Qi, X.: Pla: Language-driven
  open-vocabulary 3d scene understanding. In: CVPR (2023)

\bibitem{ding2023open}
Ding, Z., Wang, J., Tu, Z.: Open-vocabulary universal image segmentation with
  maskclip. In: ICML (2023)

\bibitem{du2022learning}
Du, Y., Wei, F., Zhang, Z., Shi, M., Gao, Y., Li, G.: Learning to prompt for
  open-vocabulary object detection with vision-language model. In: CVPR (2022)

\bibitem{geiger2012cvpr}
Geiger, A., Lenz, P., Urtasun, R.: {Are we ready for Autonomous Driving? The
  KITTI Vision Benchmark Suite}. In: CVPR (2012)

\bibitem{ghiasi2022scaling}
Ghiasi, G., Gu, X., Cui, Y., Lin, T.Y.: Scaling open-vocabulary image
  segmentation with image-level labels. In: ECCV (2022)

\bibitem{gu2021open}
Gu, X., Lin, T.Y., Kuo, W., Cui, Y.: Open-vocabulary object detection via
  vision and language knowledge distillation. ICLR  (2022)

\bibitem{ha2022semantic}
Ha, H., Song, S.: Semantic abstraction: Open-world 3d scene understanding from
  2d vision-language models. In: CoRL (2022)

\bibitem{He_2023_CVPR}
He, W., Jamonnak, S., Gou, L., Ren, L.: Clip-s4: Language-guided
  self-supervised semantic segmentation. In: CVPR (2023)

\bibitem{hegde2023clip}
Hegde, D., Valanarasu, J.M.J., Patel, V.M.: Clip goes 3d: Leveraging prompt
  tuning for language grounded 3d recognition. arXiv preprint arXiv:2303.11313
  (2023)

\bibitem{hong2021lidar}
Hong, F., Zhou, H., Zhu, X., Li, H., Liu, Z.: Lidar-based panoptic segmentation
  via dynamic shifting network. In: CVPR (2021)

\bibitem{hu2021learning}
Hu, Q., Yang, B., Xie, L., Rosa, S., Guo, Y., Wang, Z., Trigoni, N., Markham,
  A.: Learning semantic segmentation of large-scale point clouds with random
  sampling. IEEE Transactions on Pattern Analysis and Machine Intelligence
  \textbf{44}(11),  8338--8354 (2021)

\bibitem{ig20215143773}
Ilharco, G., Wortsman, M., Wightman, R., Gordon, C., Carlini, N., Taori, R.,
  Dave, A., Shankar, V., Namkoong, H., Miller, J., Hajishirzi, H., Farhadi, A.,
  Schmidt, L.: Openclip (Jul 2021). \doi{10.5281/zenodo.5143773},
  \url{https://doi.org/10.5281/zenodo.5143773}

\bibitem{jia2021scaling}
Jia, C., Yang, Y., Xia, Y., Chen, Y.T., Parekh, Z., Pham, H., Le, Q., Sung,
  Y.H., Li, Z., Duerig, T.: Scaling up visual and vision-language
  representation learning with noisy text supervision. In: ICML (2021)

\bibitem{kingma2014adam}
Kingma, D.P., Ba, J.: Adam: A method for stochastic optimization. In: ICLR
  (2015)

\bibitem{kuo2022f}
Kuo, W., Cui, Y., Gu, X., Piergiovanni, A., Angelova, A.: F-vlm:
  Open-vocabulary object detection upon frozen vision and language models. In:
  ICLR (2023)

\bibitem{lambert2020mseg}
Lambert, J., Liu, Z., Sener, O., Hays, J., Koltun, V.: Mseg: A composite
  dataset for multi-domain semantic segmentation. In: CVPR (2020)

\bibitem{li2022languagedriven}
Li, B., Weinberger, K.Q., Belongie, S., Koltun, V., Ranftl, R.: Language-driven
  semantic segmentation. In: ICLR (2022)

\bibitem{li2022panoptic}
Li, J., He, X., Wen, Y., Gao, Y., Cheng, X., Zhang, D.: Panoptic-phnet: Towards
  real-time and high-precision lidar panoptic segmentation via clustering
  pseudo heatmap. In: CVPR (2022)

\bibitem{Li_2022_CVPR}
Li, Z., Wang, W., Xie, E., Yu, Z., Anandkumar, A., Alvarez, J.M., Luo, P., Lu,
  T.: Panoptic segformer: Delving deeper into panoptic segmentation with
  transformers. In: CVPR (2022)

\bibitem{liang2023open}
Liang, F., Wu, B., Dai, X., Li, K., Zhao, Y., Zhang, H., Zhang, P., Vajda, P.,
  Marculescu, D.: Open-vocabulary semantic segmentation with mask-adapted clip.
  In: CVPR (2023)

\bibitem{Lin2017Focal}
Lin, T.Y., Goyal, P., Girshick, R., He, K., Dollar, P.: Focal loss for dense
  object detection. In: ICCV (2017)

\bibitem{liu2022open}
Liu, Q., Wen, Y., Han, J., Xu, C., Xu, H., Liang, X.: Open-world semantic
  segmentation via contrasting and clustering vision-language embedding. In:
  ECCV (2022)

\bibitem{liu2022convnet}
Liu, Z., Mao, H., Wu, C.Y., Feichtenhofer, C., Darrell, T., Xie, S.: A convnet
  for the 2020s. In: CVPR (2022)

\bibitem{loshchilov2017decoupled}
Loshchilov, I., Hutter, F.: Decoupled weight decay regularization. In: ICLR
  (2019)

\bibitem{ma2022open}
Ma, C., Yang, Y., Wang, Y., Zhang, Y., Xie, W.: Open-vocabulary semantic
  segmentation with frozen vision-language models. BMVC  (2022)

\bibitem{peng2023openscene}
Peng, S., Genova, K., Jiang, C., Tagliasacchi, A., Pollefeys, M., Funkhouser,
  T., et~al.: Openscene: 3d scene understanding with open vocabularies. In:
  CVPR (2023)

\bibitem{qi2017pointnet}
Qi, C.R., Su, H., Mo, K., Guibas, L.J.: Pointnet: Deep learning on point sets
  for 3d classification and segmentation. In: CVPR (2017)

\bibitem{qi2017pointnet++}
Qi, C.R., Yi, L., Su, H., Guibas, L.J.: Pointnet++: Deep hierarchical feature
  learning on point sets in a metric space. NeurIPS  (2017)

\bibitem{qin2023freeseg}
Qin, J., Wu, J., Yan, P., Li, M., Yuxi, R., Xiao, X., Wang, Y., Wang, R., Wen,
  S., Pan, X., et~al.: Freeseg: Unified, universal and open-vocabulary image
  segmentation. In: CVPR (2023)

\bibitem{radford2021learning}
Radford, A., Kim, J.W., Hallacy, C., Ramesh, A., Goh, G., Agarwal, S., Sastry,
  G., Askell, A., Mishkin, P., Clark, J., et~al.: Learning transferable visual
  models from natural language supervision. In: ICML (2021)

\bibitem{razani2021gp}
Razani, R., Cheng, R., Li, E., Taghavi, E., Ren, Y., Bingbing, L.: Gp-s3net:
  Graph-based panoptic sparse semantic segmentation network. In: ICCV (2021)

\bibitem{rozenberszki2022language}
Rozenberszki, D., Litany, O., Dai, A.: Language-grounded indoor 3d semantic
  segmentation in the wild. In: ECCV (2022)

\bibitem{sirohi2021efficientlps}
Sirohi, K., Mohan, R., B{\"u}scher, D., Burgard, W., Valada, A.: Efficientlps:
  Efficient lidar panoptic segmentation. IEEE Transactions on Robotics
  \textbf{38}(3),  1894--1914 (2021)

\bibitem{takmaz2023openmask3d}
Takmaz, A., Fedele, E., Sumner, R.W., Pollefeys, M., Tombari, F., Engelmann,
  F.: Openmask3d: Open-vocabulary 3d instance segmentation. In: NeuRIPS (2023)

\bibitem{tang2020searching}
Tang, H., Liu, Z., Zhao, S., Lin, Y., Lin, J., Wang, H., Han, S.: Searching
  efficient 3d architectures with sparse point-voxel convolution. In: ECCV
  (2020)

\bibitem{vaswani2017attention}
Vaswani, A., Shazeer, N., Parmar, N., Uszkoreit, J., Jones, L., Gomez, A.N.,
  Kaiser, {\L}., Polosukhin, I.: Attention is all you need. In: NeurIPS (2017)

\bibitem{wu2023pointconvformer}
Wu, W., Fuxin, L., Shan, Q.: Pointconvformer: Revenge of the point-based
  convolution. In: CVPR (2023)

\bibitem{xiao2023p3former}
Xiao, Z., Zhang, W., Wang, T., Loy, C.C., Lin, D., Pang, J.: Position-guided
  point cloud panoptic segmentation transformer. arXiv preprint  (2023)

\bibitem{xu2021rpvnet}
Xu, J., Zhang, R., Dou, J., Zhu, Y., Sun, J., Pu, S.: Rpvnet: A deep and
  efficient range-point-voxel fusion network for lidar point cloud
  segmentation. In: ICCV (2021)

\bibitem{xu2022groupvit}
Xu, J., De~Mello, S., Liu, S., Byeon, W., Breuel, T., Kautz, J., Wang, X.:
  Groupvit: Semantic segmentation emerges from text supervision. In: CVPR
  (2022)

\bibitem{xu2023open}
Xu, J., Liu, S., Vahdat, A., Byeon, W., Wang, X., De~Mello, S.: Open-vocabulary
  panoptic segmentation with text-to-image diffusion models. In: CVPR (2023)

\bibitem{xu2022simple}
Xu, M., Zhang, Z., Wei, F., Lin, Y., Cao, Y., Hu, H., Bai, X.: A simple
  baseline for open-vocabulary semantic segmentation with pre-trained
  vision-language model. In: ECCV (2022)

\bibitem{xu2022sparse}
Xu, S., Wan, R., Ye, M., Zou, X., Cao, T.: Sparse cross-scale attention network
  for efficient lidar panoptic segmentation. In: AAAI (2022)

\bibitem{yang2023regionplc}
Yang, J., Ding, R., Wang, Z., Qi, X.: Regionplc: Regional point-language
  contrastive learning for open-world 3d scene understanding. arXiv preprint
  arXiv:2304.00962  (2023)

\bibitem{yu2023fcclip}
Yu, Q., He, J., Deng, X., Shen, X., Chen, L.C.: Convolutions die hard:
  Open-vocabulary segmentation with single frozen convolutional clip. In:
  NeurIPS (2023)

\bibitem{zhang2023clip}
Zhang, J., Dong, R., Ma, K.: Clip-fo3d: Learning free open-world 3d scene
  representations from 2d dense clip. arXiv preprint arXiv:2303.04748  (2023)

\bibitem{zhou2022extract}
Zhou, C., Loy, C.C., Dai, B.: Extract free dense labels from clip. In: ECCV
  (2022)

\bibitem{zhou2023zegclip}
Zhou, Z., Lei, Y., Zhang, B., Liu, L., Liu, Y.: Zegclip: Towards adapting clip
  for zero-shot semantic segmentation. In: CVPR (2023)

\bibitem{Zhou2021PanopticPolarNet}
Zhou, Z., Zhang, Y., Foroosh, H.: Panoptic-polarnet: Proposal-free lidar point
  cloud panoptic segmentation. In: CVPR (2021)

\bibitem{zou2023generalized}
Zou, X., Dou, Z.Y., Yang, J., Gan, Z., Li, L., Li, C., Dai, X., Behl, H., Wang,
  J., Yuan, L., et~al.: Generalized decoding for pixel, image, and language.
  In: CVPR (2023)

\end{thebibliography}
\appendix
\clearpage

\section{PFC Baseline}
\label{sec:s1}
\begin{figure*}
\begin{center}
\vspace{-15pt}
\includegraphics[width=\linewidth]{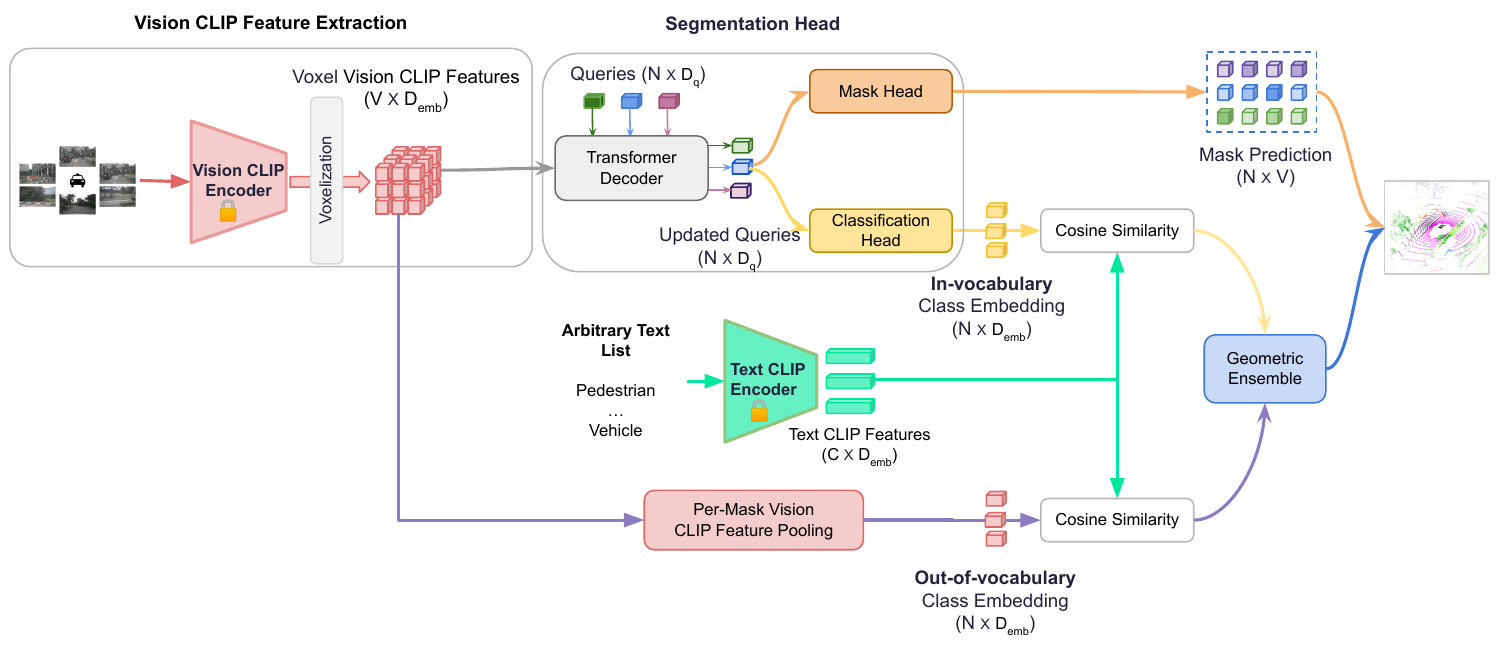}
\end{center}
\vspace{-15pt}
\caption{The overview of PFC baseline. The PFC baseline takes frozen voxel vision CLIP features as inputs. We modified the segmentation head of a closed-set model so that it can predict in-vocabulary class embedding. The classification score is predicted by computing the cosin similarity between the predicted embedding and the text CLIP embedding. During testing, we further apply per-mask vision feature pooling to obtain out-of-vocabulary class embedding. The final per-mask classification logits are the geometric ensemble of in-vocabulary and out-of-vocabulary classification results.}
\label{fig:baseline_fig}
\end{figure*}
As there is no existing work for the 3D open vocabulary panoptic segmentation task, a natural idea would be to extend the 2D state-of-the-art open vocabulary segmentation to 3D. We start by adapting the essential components from FC-CLIP~\cite{yu2023fcclip}, a 2D open-vocabulary segmentation model that achieves state-of-the-art performances across different datasets, to P3Former~\cite{xiao2023p3former}, a state-of-the-art 3D closed-set panoptic segmentation model. Since the models of FC-CLIP and P3Former are very different, we conduct some necessary changes to the architecture of P3Former. We name this baseline PCFormer+FC-CLIP (PFC). The overall architecture of the PFC is shown in~\cref{fig:baseline_fig}.

\xhdr{Vision CLIP feature extraction.} FC-CLIP\cite{yu2023fcclip} demonstrates that frozen CLIP features can produce promising classification performance on both base and novel classes. In the same spirit, we construct a Vision CLIP feature extractor as follows: a pre-trained V-L segmentation model~\cite{ghiasi2022scaling} is applied to extract pixel-wise CLIP features from each camera image. Within each voxel, every LiDAR point is projected into its corresponding camera image based on the intrinsic and extrinsic calibration parameters, in order to index into the corresponding vision CLIP features. The vision CLIP features of all the points belonging to each voxel are then averaged to represent that voxel. The voxel CLIP features will be referred as $F_{vclip} \in \mathbb{R}^{V\times D_{emb}}$, where $V$ is the number of voxels after voxelization and $D_{emb}$ is the dimension of the CLIP features. Note that the Vision CLIP encoder is frozen and it is identical to the one in our proposed method.

\xhdr{Segmentation head.} We use one learnable querie $q$ to represent each instance or thing. Queries matched with groudtruth objects are supervised with both classification loss and mask loss. FC-CLIP~\cite{yu2023fcclip} shows that the mask generation is class-agnostic, and therefore we follow FC-CLIP and only modify the classification head to add a class embedding. Specifically, the class embedding $f_{cls}$ prediction is defined as: 
\begin{equation}
    v_{q} = f_{cls}(q) \in{\mathbb{R}^{D_{emb}}},
\label{eq:vq_s}    
\end{equation}
where $v_{q}$ is in the CLIP embedding space. The predicted class logits are then computed from the cosine similarity between the predicted class embedding and the text embedding of every category name from the evaluation set using a frozen CLIP model. The classification logits are defined as:
\begin{equation}
    s_{v_{q}} = \frac{1}{T}[\cos(v_{q},t_1),\cos(v_{q},t_2), \dots, \cos(v_{q},t_C)]
\label{eq:svq_sup}
\end{equation}
where $t_i \in{\mathbb{R}^{D_{emb}}}$, $i\in\{1,2,\dots,C\}$ is the text embedding, $C$ is the number of categories ($C_B$ in training and $C_B$+$C_N$ in testing), and $T$ is a learnable temperature term that controls the concentration of the distribution. Following FC-CLIP, we name this trainable classifier the \textbf{in-vocabulary} classifier. The loss function, then, is $L= w_\alpha * L_{cls}+ w_\beta * L_{mask}$, where $L_{cls}$ and $L_{mask}$ are the softmax cross-entropy classification loss and mask loss, respectively. $w_\alpha$ and $w_\beta$ are weights for classification loss and mask loss, respectively. Note that, for classification, we apply a softmax cross-entropy loss instead of focal loss because of the following ensembling process.

\xhdr{Geometric ensemble.} Previous open-vocabulary works~\cite{gu2021open,ghiasi2022scaling,kuo2022f,xu2023open,yu2023fcclip} show that a trainable in-vocabulary classifier fails to make good predictions for novel classes. During testing, we follow \cite{xu2023open,yu2023fcclip}, and construct an \textbf{out-of-vocabulary} classifier that utilizes voxel Vision CLIP features to get an embedding for each query $q$ by mask pooling the Vision CLIP features:
\begin{align}\label{eq:mask_pooling}
    w_{q} =& \frac{1}{|M_q|}\sum_{p} \mathds{1}(p\in M_q) F_{vclip}(p)
\end{align}
, where $M_q$ is the set of points, $p$, belonging to the mask for query, $q$.
The out-of-vocabulary classification logits $s_{w_q}$can be computed as
\begin{equation}
    s_{w_{q}} = \frac{1}{T}[\cos(v_{q},t_1),\cos(v_{q},t_2), \dots, \cos(v_{q},t_C)]
\label{eq:swq_sup}
\end{equation}
where the temperature term $T$ is the same is the one in~\cref{eq:svq_sup}. Note that the out-of-vocabulary classifier is frozen and is only applied during testing.
The final classification score is computed as the geometric ensemble of the in-vocabulary classifier and out-of-vocabulary classifier for every class, $i$:
\begin{equation}
s_{{g}_{q}}(i) = \begin{cases}
p_{v_{q}}(i)^{1-\alpha} p_{w_{q}}(i)^{\alpha} &\text{if } i \text{$\in C_{B}$}\\
p_{v_{q}}(i)^{1-\beta} p_{w_{q}}(i)^{\beta} &\text{if } i \text{$\in C_{N}$}
\end{cases}
\end{equation}
where $p_{v_{q}} = \text{softmax}(s_{v_{q}} )$, $p_{w_{q}} = \text{softmax}(s_{w_{q}} )$ are the derived probabilities and $\alpha,\beta \in[0,1]$ are hyperparameters to control the contributions of in-vocabulary classifier and out-of-vocabulary classifier. In practice, we try multiple pairs of $\alpha,\beta$ and report the result of the best pair. We have found that $\alpha=0$ and $\beta=1$ generates the best results for the PFC baseline in all different base/novel splits, which indicates that the baseline solely relies on out-of-vocabulary classifier to make predictions for novel classes.

\section{Query Assignment}
\label{sec:s2}
\begin{figure*}
\begin{center}
\vspace{-15pt}
\includegraphics[width=\linewidth]{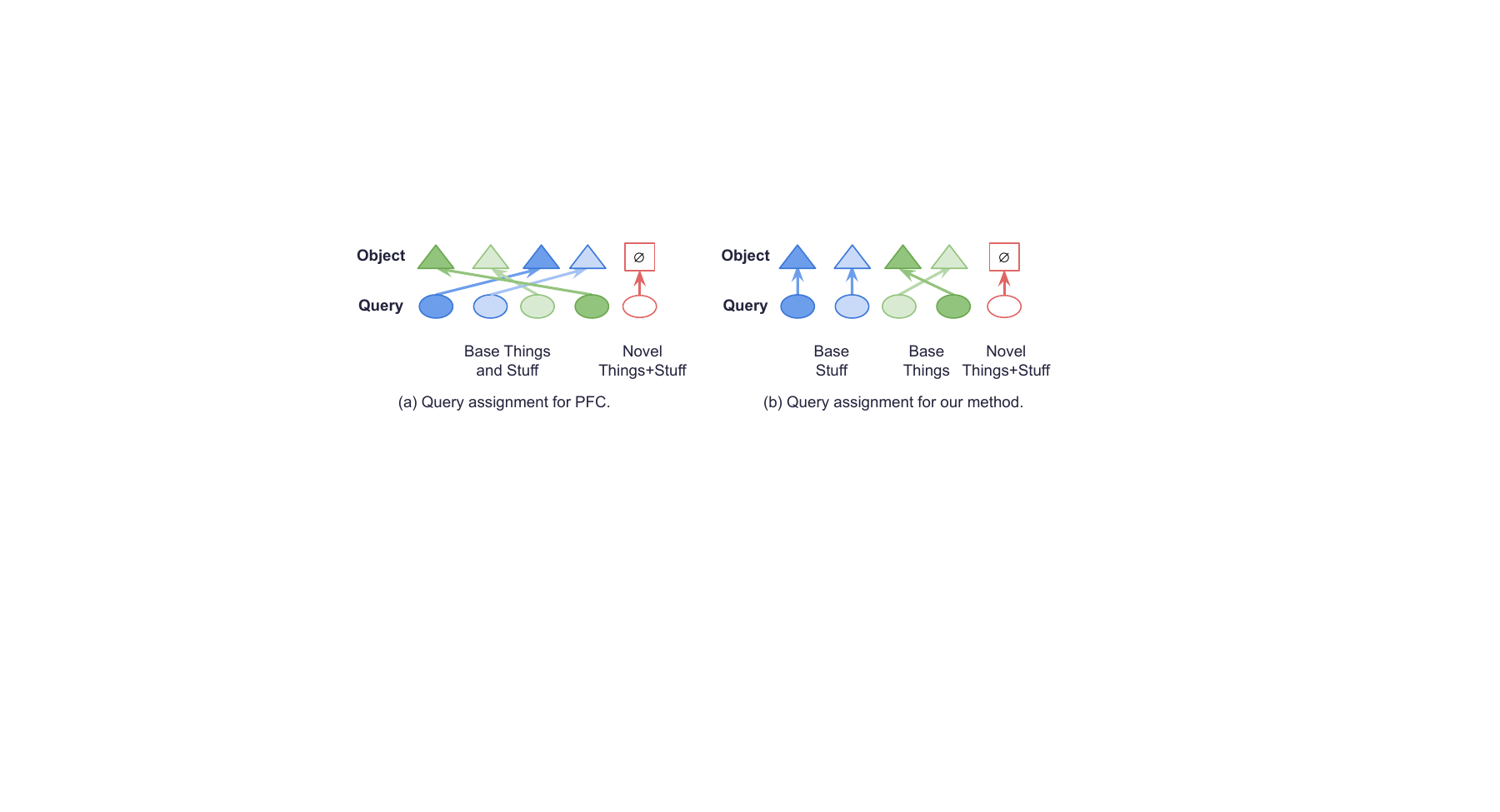}
\end{center}
\vspace{-15pt}
\caption{Visualization for the two strategies for query assignment. 
}
\label{fig:query_assignment_fig}
\end{figure*}

For both the baseline and our method, a single query is used to represent an individual object. This requires specific query assignment strategies to match predictions with groundtruth base objects during training. FC-CLIP uses one set of learnable queries to make predictions for base and novel classes. Therefore, the same set of queries are matched with base \textit{thing} and \textit{stuff} objects. The unmatched queries are potentially in charge of making predictions for novel \textit{thing} and \textit{stuff} objects, as shown in~\cref{fig:query_assignment_fig} (a). In contrast, our method uses two sets of queries. The first query set is used to represent base \textit{things} classes after bipartite matching, while the second, fixed, query set is for base \textit{stuff} classes, as shown in~\cref{fig:query_assignment_fig} (b). The separation of base \textit{things} queries and base \textit{stuff} queries makes our model converge faster and improves overall performance.

\section{Intuition of Voxel-level Distillation Loss}
\label{sec:s3}
\begin{figure*}
\begin{center}

\includegraphics[width=\linewidth]{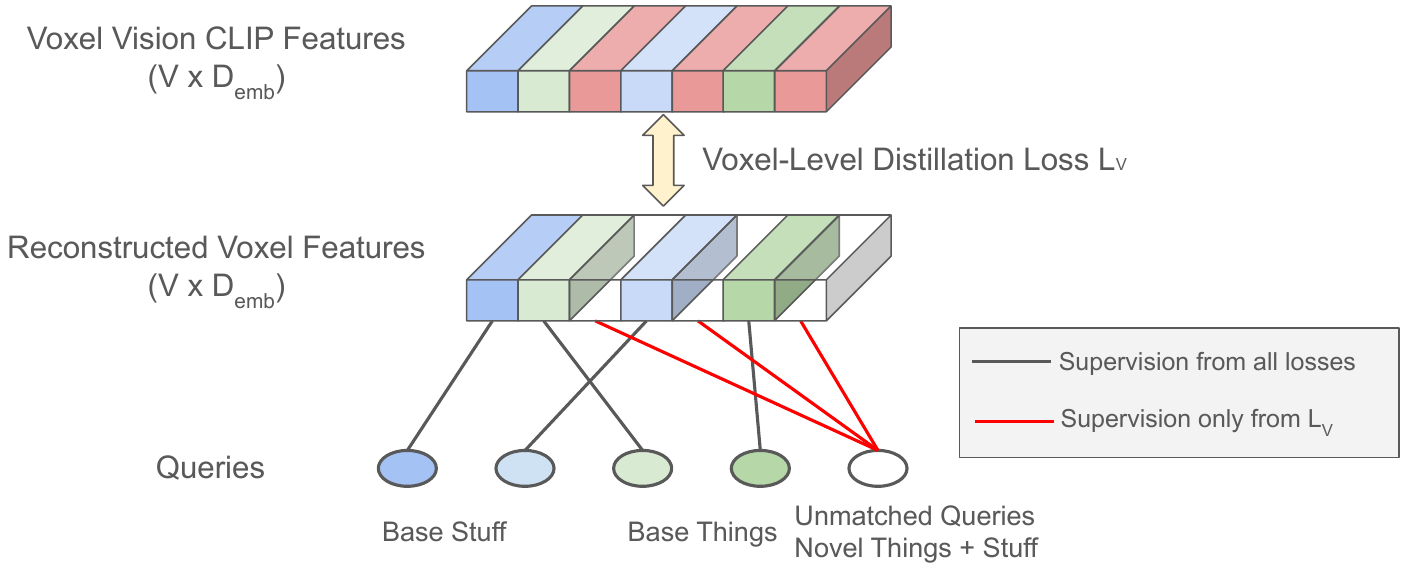}
\end{center}
\vspace{-15pt}
\caption{Intuition of the voxel-level distillation loss ($L_V$). $L_V$ does not require any labels during training and can be optimized on all queries and voxeles. Therefore, it encourages the unmatched queries to make predictions in the voxels that have no supervision from the base classes. 
}

\label{fig:voxel_loss}
\end{figure*}


\cref{fig:voxel_loss} illustrates the intuition behind the proposed voxel-level distillation loss $L_V$. $L_V$ is not dependent on any labels, and therefore, can be applied to all the queries. For voxels belonging to base classes, all loss functions will be enforced during optimization, including the two standard loss functions (per-query mask classification loss $L_{cls}$ and per-query mask prediction loss $L_{mask}$), 
the proposed object-level distillation loss $L_O$ and the proposed voxel-level distillation loss $L_V$. For voxels belonging to queries that do not match to any labels, likely being novel \textit{things} or novel \textit{stuff} objects, they will be mainly supervised by the proposed voxel-level distillation loss $L_V$. With the help of this loss, the unmatched queries learn to make predictions in the voxels with no supervision from the base classes. In this way, we enforce the supervision on all the queries and voxels and the model can learn to produce meaningful predictions for both base and novel categories.

\section{
More Experimental Results} \label{sec:s4}
\xhdr{Queries in object-level distillation loss.} In our loss function design for the object-level distillation loss $L_O$, we only enforce constraints on queries matched with base classes. One natural question would be: can we apply the constraint on all queries to improve predictions? We conduct an ablation study for this, with results shown in~\cref{table:ns_sup_ob}. We consider $PQ$ as the most important metric. When we apply the object-level distillation loss to all queries, the overall performance is slightly worse, especially for the novel \textit{stuff} classes.

\begin{table*}[t]

\small
\centering
\caption{\textbf{Impact of queries in $L_O$}. We conduct an ablation study comparing applying $L_O$ to matched queries vs all queries. The overall performance is better when we only apply $L_O$ on matched queries.}
\begin{tabular}{c|c|c|c|c|c|c|c|c|c|c}
\specialrule{.2em}{.1em}{.1em}
 Queries in $L_O$  & $PQ$  & $ PQ_{N}^{Th} $ & $ PQ_{N}^{St} $  & $RQ$ & $ RQ_{N}^{Th} $ & $ RQ_{N}^{St} $ & $SQ$  & $ SQ_{N}^{Th} $ & $ SQ_{N}^{St} $ & mIoU\\ 
\hline
Matched Only          & \textbf{62.0}                    & 49.6                  & \textbf{35.2}                  & \textbf{70.9}                  & 55.6                 & \textbf{46.0}                  & \textbf{87.0}                    & \textbf{89.1}                  & 76.7  &  60.1 \\ \hline
All                   & 61.0                   & \textbf{49.9}                  & 25.4                  & 70.0                 & \textbf{56.3}                 & 34.4                  & 86.3                    & 74.3                 & \textbf{88.7} &  \textbf{60.5} \\ 
\hline
\end{tabular}

\label{table:ns_sup_ob}
\end{table*}

\begin{table*}[t]
\small
\centering
\caption{\textbf{Performance of panoptic segmentation on nuScenes with a different split}. We compare the performance with a different split with 4 novel classes (B12/N4). The novel \textit{things} classes are construction vehicle and traffic cone. The novel \textit{stuff} classes are other-flat and man-made. Our method consistently outperforms the PFC baseline across almost all the metrics by a large margin.}
\vspace{-2mm}
\renewcommand{\arraystretch}{1.0}
	\vspace{2mm}
	\newcommand{\tabincell}[2]{\begin{tabular}{@{}#1@{}}#2\end{tabular}}
	\centering
	\renewcommand{\arraystretch}{1.43}
	\resizebox{\textwidth}{!}
	{
		\centering
        \begin{tabular}{c|c|c|c|c|c|c|c|c|c|c|c|c}
        \specialrule{.2em}{.1em}{.1em}
        Model & Type & Supervision  & $PQ$  & $ PQ_{N}^{Th} $ & $ PQ_{N}^{St} $  & $RQ$ & $ RQ_{N}^{Th} $ & $ RQ_{N}^{St} $ & $SQ$  & $ SQ_{N}^{Th} $ & $ SQ_{N}^{St} $ & mIoU\\ 
        \hline
        P3Former\cite{xiao2023p3former} &  closed-set  & full     & 75.8   & 76.4 & 86.9 & 83.8  & 84.8  & 98.3 & 90.1   & 89.8 & 88.4 &  78.2\\ 
        \hline
        \hline
        PFC &  open-voc  & partial 	& 49.9	& 22.5	& 14.0	& 59.6	& \textbf{26.9}	& 21.9	& 85.6 & 82.9 & 61.0 & 53.8 \\ \hline
        Ours &  open-voc  & partial                   & \textbf{55.4}                    & \textbf{23.2}                  & \textbf{24.7}                  & \textbf{62.9}                  & 26.0                 & \textbf{29.8}                  & \textbf{85.6}                    & \textbf{87.6}                 & \textbf{69.3}  &  \textbf{55.0} \\ 
        \hline
        \end{tabular}
        }
\label{table:ns_sup_2}
\end{table*}

\begin{table*}[t]
\small
\centering
\caption{\textbf{Performance of panoptic segmentation on nuScenes with a different split}. We compare the performance with a different split with 4 novel classes (B12/N4). The novel \textit{things} classes are construction vehicle and traffic cone. The novel \textit{stuff} classes are other-flat and man-made. Our method consistently outperforms the PFC baseline across almost all the metrics by a large margin.}
\vspace{-2mm}
\renewcommand{\arraystretch}{1.0}
	\vspace{2mm}
	\newcommand{\tabincell}[2]{\begin{tabular}{@{}#1@{}}#2\end{tabular}}
	\centering
	\renewcommand{\arraystretch}{1.43}
	\resizebox{\textwidth}{!}
	{
		\centering
        \begin{tabular}{c|c|c|c|c|c|c|c|c|c|c|c|c}
        \specialrule{.2em}{.1em}{.1em}
        Model & Type & Supervision  & $PQ$  & $ PQ_{N}^{Th} $ & $ PQ_{N}^{St} $  & $RQ$ & $ RQ_{N}^{Th} $ & $ RQ_{N}^{St} $ & $SQ$  & $ SQ_{N}^{Th} $ & $ SQ_{N}^{St} $ & mIoU\\ 
        \hline
        P3Former\cite{xiao2023p3former} &  closed-set  & full     & 75.8   & 71.1 & 96.2 & 83.8  & 78.8  & 99.9 & 90.1   & 90.1 & 96.3 &  78.2\\ 
        \hline
        \hline
        PFC &  open-voc  & partial 	& 43.1	& 16.4	& 2.1	& 51.6	& 20.0	& 3.0	& 79.8 & 83.8 & \textbf{69.7} & 43.5 \\ \hline
        Ours &  open-voc  & partial                   & \textbf{53.1}                    & \textbf{31.0}                  & \textbf{35.1}                  & \textbf{63.0}                  & \textbf{35.2}                & \textbf{53.8}                  & \textbf{82.3}                    & \textbf{87.7}                 & 65.3  &  \textbf{50.5} \\ 
        \hline
        \end{tabular}
        }
\label{table:ns_sup_3}
\end{table*}

\begin{table}[t]
\small
\centering
\caption{\textbf{Performance on novel \textit{stuff} classes}. We compare the performance of PFC, our method and the oracle setting that based on the GT masks on novel \textit{stuff} classes. The spilits are the same as in Tab. 1 of the main paper.}
\begin{tabular}{c|c|c|c|c}
\specialrule{.2em}{.1em}{.1em}
\multicolumn{1}{l|}{} & mIoU  & PQ    & RQ    & SQ    \\ \hline
PFC                    & 4.38  & 0.5   & 0.83  & 60.44 \\ \hline
Ours                & 45.14 & 35.25 & 45.97 & 76.69 \\ \hline
Oracle (GT Masks)   & 51.39 & 52.61 & 53.00    & 99.26 \\ \hline
\end{tabular}
\vspace{-3mm}

\label{table:ns_sup_4}
\end{table}

\xhdr{More splits.} In order to show that our proposed method generalizes well in different scenarios, we conduct experiments on two more random B12/N4 splits.
As shown in~\cref{table:ns_sup_2} and~\cref{table:ns_sup_3}, our method surpasses the PFC baseline in almost all metrics across all the splits, demonstrating the capability of our proposed method.

{
\xhdr{Performance on novel \textit{stuff} classes.} The performance of PFC baseline is almost $0$ on novel \textit{stuff} classes. To verify whether it is due to poor mask predictions for the novel stuff calss, we conduct an oracle experiment by max-pooling vision CLIP features with ground truth masks and then use its similarity with CLIP text features to determine its category, and the results are shown in~\cref{table:ns_sup_4}. We achieve $53$ $RQ$ using ground truth mask, which demonstrate that the bad performance of PFC baseline is indeed due to poor mask quality. Also, the low RQ number shows that the prediction task on novel stuff class is very challenging.
}

\section{Discussion} \label{sec:s5}
\xhdr{Class-agnostic mask generator.} As shown in FC-CLIP~\cite{yu2023fcclip}, the mask head is class-agnostic if we do not apply any penalty to unmatched queries. We follow the same strategy in our paper. The metrics $SQ_N^{Th}$ and $SQ_N^{St}$ in all experiments indicate that the mask predictions for both \textit{things} and \textit{stuff} are reasonable.

\xhdr{Comparison with RegionPLC.} RegionPLC~\cite{yang2023regionplc} proposes to take advantage of regional visual prompts to create dense captions. After point-discriminative contrastive learning, the model can be used for semantic segmentation or instance segmentation. There are two main differences between RegionPLC and our method: 1. RegionPLC addresses the problem of semantic segmentation or instance segmentation individually, while our model addresses semantic segmentation and instance segmentation in the same model. 2. RegionPLC focuses on getting point-level discriminative features, while our model takes the pretrained CLIP features as input and aims to build model architecture and design loss functions. In our method, we do not compare with RegionPLC because the experiment settings are different and there is no public code to reproduce the contrastive learning process. However, we do think there is great potential in combining RegionPLC and our method. One idea would be to replace the vision CLIP features in our model with the features derived from RegionPLC.
\section{Visualization}
\label{sec:s6}
\begin{figure*}[!hbt]
\begin{center}
\includegraphics[width=1.0\linewidth]{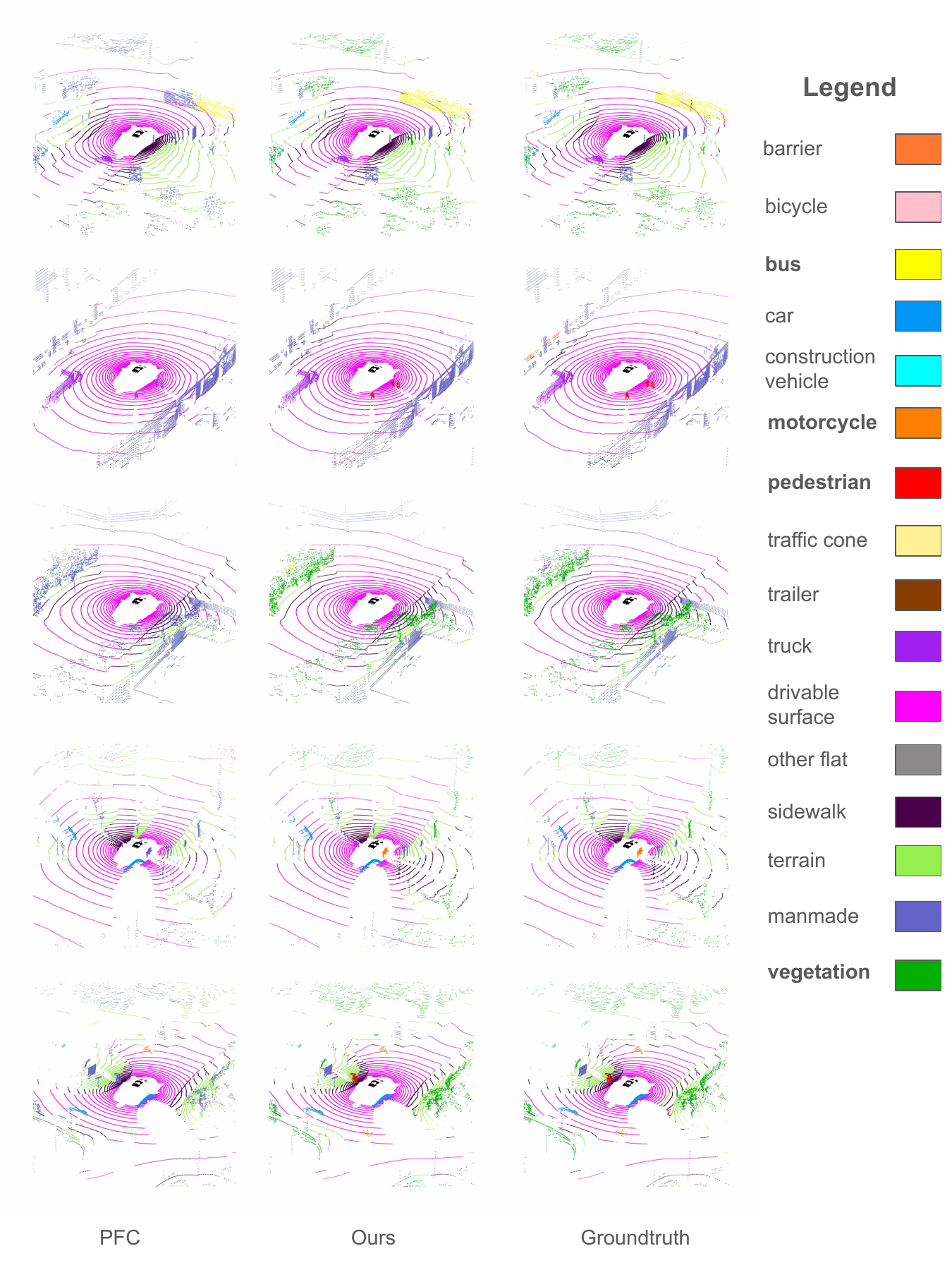}
\end{center}
\caption{Qualitative Results in nuScenes Dataset. We present the comparison among PFC, our method and the groundtrth. The novel objects are marked in \textbf{bold} in the legend. 
}
\label{fig:vis_nuscenes}
\end{figure*}

\begin{figure*}[!htb]
\begin{center}
\includegraphics[width=1.0\linewidth]{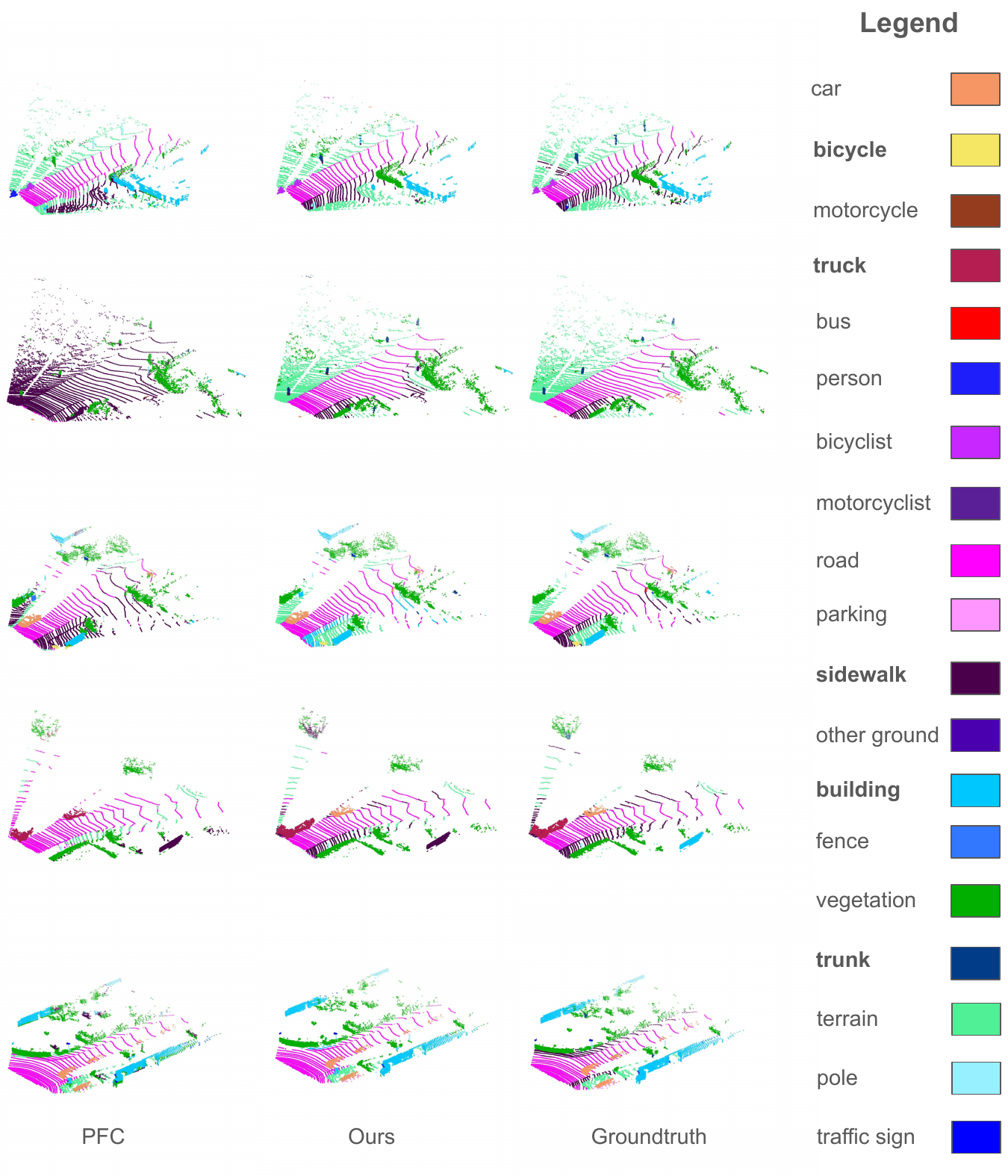}
\end{center}
\caption{Qualitative Results in SemanticKITTI Dataset. We present the comparison among PFC, our method and the groundtrth. The novel objects are marked in \textbf{bold} in the legend.
}
\label{fig:vis_sk}
\end{figure*}

We present the visualization of PFC baseline, our method and groundtruth in~\cref{fig:vis_nuscenes} and~\cref{fig:vis_sk}. Note that we only visualize the points that are visible in frontal camera views in~\cref{fig:vis_sk}.

\end{document}